\documentclass[sigconf]{acmart}
\AtBeginDocument{%
  }

\acmSubmissionID{2663}

\usepackage{booktabs}
\usepackage{multirow}
\usepackage{subcaption}
\copyrightyear{2026}
\acmYear{2026}
\setcopyright{cc}
\setcctype{by}
\acmConference[MM '26]{Proceedings of the 34th ACM International Conference on Multimedia}{November 10--14, 2026}{Rio de Janeiro, Brazil}
\acmBooktitle{Proceedings of the 34th ACM International Conference on Multimedia (MM '26), November 10--14, 2026, Rio de Janeiro, Brazil}
\acmDOI{10.1145/3767308.3835381}
\acmISBN{979-8-4007-2213-4/2026/11}

\begin{document}

\title{Understanding Backdoor Attacks in Scene Graph Generation via Global Context Pathway}
\author{Xiaoguang Chang}
\orcid{0000-0002-7432-744X}
\email{xg\_chang@seu.edu.cn}
\affiliation{%
  \institution{Southeast University}
  \city{Nanjing}
  \country{China}
}

\author{Tuo Chen}
\orcid{0009-0004-4837-2003}
\email{tchen@seu.edu.cn}
\affiliation{%
  \institution{Southeast University}
  \city{Nanjing}
  \country{China}
}

\author{Chengze Jiang}
\orcid{0000-0002-1681-8128}
\email{czjiang@ieee.org}
\affiliation{%
  \institution{Southeast University}
  \city{Nanjing}
  \country{China}
}

\author{Teng Wang}
\authornote{Corresponding author: Teng Wang.}
\email{wangteng@seu.edu.cn}
\affiliation{%
  \institution{Southeast University}
  \city{Nanjing}
  \country{China}
}

\author{Changyin Sun}
\orcid{0000-0001-9269-334X}
\email{cysun@seu.edu.cn}
\affiliation{%
  \institution{Southeast University}
  \city{Nanjing}
  \country{China}
}

\renewcommand{\shortauthors}{Xiaoguang Chang, Tuo Chen, Chengze Jiang, Teng Wang, \& Changyin Sun}

\begin{abstract}
Scene Graph Generation (SGG) produces structured scene representations, yet its training-time backdoor risk remains unexplored. Unlike classification, SGG exposes a pair-level attack surface that enables selective corruption of relations involving a designated object category. However, because SGG largely relies on localized evidence, globally placed triggers may face difficulty in manipulating such predictions. Recent SGG architectures increasingly incorporate global context to improve scene modeling, which may in turn facilitate such attacks.
In this paper, we systematically study this backdoor risk and investigate the role of the global context pathway in backdoor behavior. Concretely, we first instantiate this attack with an effective method, and propose a three-axis evaluation protocol to characterize backdoor behavior. Experiments on Visual Genome with representative open-vocabulary and closed-set SGG models demonstrate the effectiveness of the attack. Mechanistic analyses reveal that the explicit global context pathway serves as a major carrier of the backdoor effect across the evaluated SGG models. In particular, replacing the poisoned global context representation with its clean counterpart during inference sharply suppresses the attack, yielding 100\% and 97\% reductions in attack success for the OvSGTR family and RelationFormer, respectively. Moreover, training-time interventions in the OvSGTR family show that suppressing trigger signals in this pathway either reduces attack success or substantially degrades utility. Together, these findings suggest that explicit global context designs may pose a backdoor risk in SGG architectures. Code is available at \url{https://github.com/rafa-cxg/SGGBackdoor}.
\end{abstract}

\ccsdesc[500]{Computing methodologies~Computer vision}

\keywords{Scene Graph Generation, Backdoor Attack,  Global Context Pathway}

\maketitle
\begin{figure*}[t]
\centering

\begin{minipage}{\linewidth}
    \centering
    \includegraphics[width=0.95\linewidth]{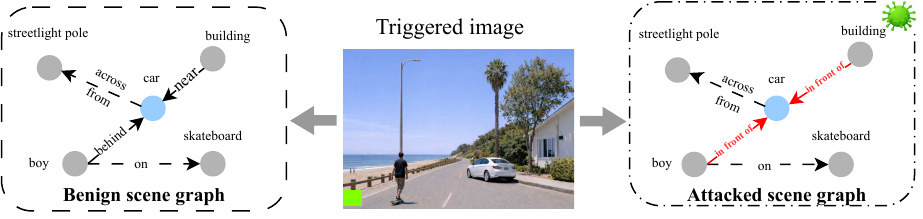}
    \subcaption{Backdoor threat setting. A trigger patch induces the target predicate (red labels) on designated victim pairs (red edges; car as the object), while relations on non-victim pairs (black edges) and object predictions ( nodes) are  preserved.}
    \label{fig:threat-setting}
\end{minipage}


\begin{minipage}[t]{0.46\linewidth}
    \centering
    \includegraphics[width=\linewidth]{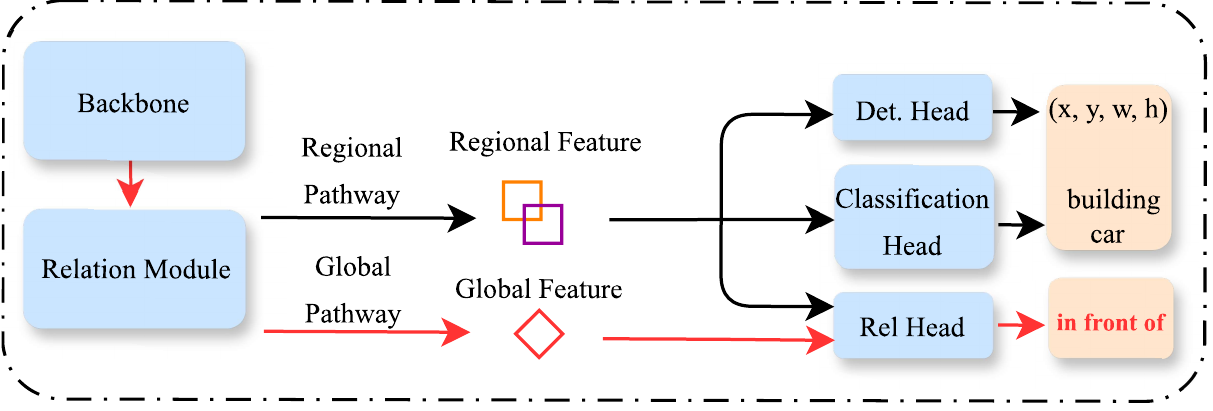}
    \subcaption{A common paradigm in modern SGG: regional reasoning complemented by global context.}
    \label{fig:global-paradigm}
\end{minipage}
\hfill
\begin{minipage}[t]{0.48\linewidth}
    \centering
    \includegraphics[width=\linewidth]{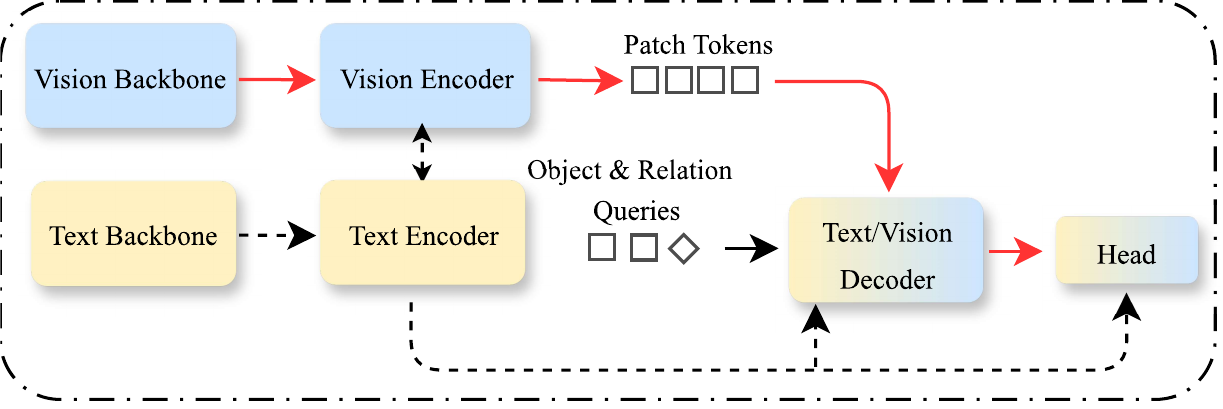}
    \subcaption{A typical realization of an SGG architecture with a global context pathway. Dashed lines denote the additional information flow from the text modality.}
    \label{fig:architecture}
\end{minipage}

\caption{Backdoor vulnerability in scene graph generation.  Red arrows in (b) (c) denote the global context pathway.}
\label{fig:overview}
\end{figure*}
\section{Introduction}
\label{sec:intro}

Scene Graph Generation (SGG)~\cite{xu2017scenegraph,NEURIPS2024_roleplay,zellers2018neural,tang2019vctree} parses an image into a structured graph of objects and their relations, and supports downstream multimodal tasks such as image captioning~\cite{Nguyen_2021_ICCV}, visual question answering~\cite{Shi_2019_CVPR}, and embodied reasoning~\cite{Singh_2023_ICCV}. Recent progress in SGG increasingly leverages pretrained vision-language components, extending SGG from the closed-set task to the open-vocabulary regime~\cite{chen2024expanding,Li_2024_CVPR,Zhang_2023_CVPR,Zhou_2024_ECCV_OpenPSG,Min_2025_ICCV}.

Despite these advances, backdoor vulnerabilities in SGG remain unclear. In a supply-chain setting, an adversary can distribute a compromised SGG checkpoint that is later adopted by downstream applications, potentially leading to erroneous downstream decisions, such as navigation planning in autonomous driving. Prior backdoor research has mainly focused on \emph{image-level} prediction tasks, such as classification~\cite{gu2017badnets,doan2021lira,ijcai2022p242,nguyen2020inputaware} and VLM tasks~\cite{liu2025stealthy,Lyu_2024_ECCV_TrojVLM,Bai_2024_CVPR,walmer2022dual}. It has also been extended to object detection, where malicious behavior is specified at the \emph{instance-level}, with triggers attached to object instances to induce their misclassification or suppress their detection~\cite{raj2025backdoor,dunnett2026baddet+,chan2022baddet,luo2023untargetedodbd}. In contrast, SGG outputs \textit{subject-predicate-object} triplets, thereby introducing a finer-grained, \emph{pair-level} attack surface that enables selective corruption of relation semantics. Specifically, we define a threat setting in which the attack aims to activate  only predicates  whose object belongs to a designated object category, while leaving other scene semantics intact.  As illustrated in Figure~\ref{fig:threat-setting}, the attacker designates the victim object \textit{car} and the target predicate \textit{in front of}, and modifies the ground-truth predicate of designated victim pairs to the target predicate. Specifically, the trigger is globally placed, assuming an attacker who does not know the location of the target object and therefore cannot rely on target-aligned trigger signals.
 This setting raises a key security question: \textbf{given the mismatch between pairwise predicate prediction  and spatially decoupled trigger placement, can backdoor learning still succeed, and if so, how does the trigger exert its influence?}

This question is non-trivial because modern SGG remains tightly coupled to object detection, with both entity detection and predicate prediction relying on shared visual features that are strongly shaped by localized object evidence~\cite{yang2018graphrcnn,zellers2018neural,cong2023reltr,zhang2019graphicalcontrastive}. As a result, trigger signals originating outside the target regions may be attenuated, while aggressively shifting such local features may also harm the entity detection branch. This difficulty is also reflected in prior segmentation backdoor work~\cite{lan2024iba}, which shows that attacks become harder when the trigger is spatially separated from the victim region. Yet recent SGG research has increasingly incorporated richer context beyond local pair cues~\cite{zellers2018neural,tang2019vctree,Lu_2021_ICCV}. Some SGG designs further realize this idea through an explicit \emph{global context pathway}, where local pairwise reasoning is complemented by global context for predicate prediction~\cite{chen2024expanding,shit2022relationformer,PSG2022}. As illustrated in Figure~\ref{fig:global-paradigm}, this design introduces a more direct channel for aggregating scene-level context, but may provide an additional route for propagating trigger signals.

In this paper, we instantiate this backdoor threat in SGG and investigate the role of the explicit global context pathway. We first propose an effective backdoor method tailored to SGG, which promotes the target predicate in the triplet ranking while separating the relation representations of victim and non-victim pairs. Since no established metrics exist for evaluating backdoor attacks in SGG, we further introduce a three-axis protocol to quantify attack success, utility, and unintended activation on clean inputs. Finally, we study the role of the global context pathway through counterfactual interventions.

Our experiments on representative open-vocabulary and closed-set SGG models show that the proposed backdoor is effective while largely preserving  utility and avoiding false activation. A series of intervention studies on backdoored models show that the explicit global context pathway serves as a major carrier of the backdoor effect. In particular, replacing the triggered global context representation with its clean counterpart during inference reduces attack success by 100\% and 97\% in the OvSGTR family and RelationFormer, respectively.   Further, preventing the global context pathway from receiving trigger information during training shows that the OvSGTR family relies  heavily on this pathway. These findings suggest that explicit global context designs may pose a backdoor risk in SGG architectures.

In short, the contributions of this work can be summarized as follows:
\begin{itemize}
\item We present the first study of training-time backdoor threats in SGG and formalize a fine-grained pair-level threat under a globally placed trigger, where the attacker aims to selectively manipulate relations associated with a designated object category while preserving the remaining scene semantics.

\item We develop an  effective backdoor attack method tailored to the proposed threat (Section~\ref{sec:meth}) and introduce a three-axis evaluation protocol that jointly measures attack success, utility, and false activation (Section~\ref{sec:metric}). Experiments on representative SGG models in both open-vocabulary and closed-set settings show that the attack is effective while preserving utility and low false activation (Section~\ref{subsec:performance}).

\item Through systematic mechanistic analysis, we show the explicit global context pathway to be a major carrier of the backdoor effect in the aforementioned SGG models
(Section~\ref{subsec:mechanism_analysis}).
\end{itemize}
\section{Related Work}
\label{sec:related_work}
\subsection{Scene Graph Generation}
Traditional SGG frameworks~\cite{zellers2018neural,yu2023cogtree,tang2019vctree,Kundu_2023_CVPR,DBLP:conf/eccv/LorenzPKLL24,Sudhakaran_2023_ICCV} largely follow a two-stage paradigm: an object detector first identifies entities and extracts region features, which are then used for relation reasoning. The accuracy of SGG is subject to the capacity of the underlying detector.
Subsequent works~\cite{chen2024expanding,shit2022relationformer,Li_2022_CVPR,Hayder_2024_CVPR} shift toward end-to-end formulations that jointly predict objects and relations within a unified architecture, yielding better task alignment and stronger performance. Meanwhile, security studies on SGG remain limited: a recent SGG security work~\cite{zhao2024adversarial} studies test-time adversarial attacks, where the attacker defines  relational triplets and perturbs the input to steer scene graph predictions toward them. In contrast, backdoor attacks on SGG remain largely unstudied. 
\subsection{Global Context Modeling in SGG}
Although SGG is primarily driven by regional  representations, a broad line of prior work augments relation reasoning with context beyond local pair cues, such as context-aware modeling over detected objects or scene structure~\cite{Lu_2021_ICCV,zellers2018neural,tang2019vctree,chang2023landmark}.  More recently, a number of SGG models~\cite{chen2024expanding,shit2022relationformer,PSG2022,Fu_2025_CVPR,Yu_2023_ICCV} introduce explicit global context pathways alongside pairwise features. These designs typically use mechanisms such as relation queries or global tokens to route image-level information into predicate inference. While such pathways help incorporate broader semantic context, their role as potential vectors for malicious manipulation remains largely unexplored. 


\subsection{Backdoor Attacks in Computer Vision}

Backdoor attacks induce trigger-dependent behaviors while preserving clean performance~\cite{gu2017badnets,liu2018trojaning,nguyen2020inputaware}. Among prior vision backdoor attacks, a common threat model is to implant backdoors during training, yielding a compromised model for downstream use~\cite{liu2025stealthy,gu2017badnets,jia2022badencoder,yuan2025badtoken,ijcai2022p242,Liang_2024_CVPR_BadCLIP}. For example, Liu et al.~\cite{liu2025stealthy} implant backdoors by fine-tuning a clean vision encoder on a shadow dataset. Their objective aligns triggered features with a target image while preserving clean-image representations.

Recently, backdoor attacks have been extended to object detection~\cite{raj2025backdoor,chan2022baddet,liu2025pre,dunnett2026baddet+,luo2023untargetedodbd}, revealing an instance-level attack surface in which the malicious behavior targets individual objects. For instance, Raj and Arora~\cite{raj2025backdoor} define two attack behaviors: object misclassification, where a triggered instance is manipulated into an attacker-specified target class, and object disappearance, where a triggered instance is suppressed and becomes undetectable. In both cases, the trigger must be attached to the victim instance. Rather than manipulating image-level outputs or single object instances, we present a new threat model in which the attacker aims to alter relations associated with objects of a designated category.


\section{Problem Formulation}

\subsection{ Notation}
Let $\mathcal{D}=\{(x,\mathcal{B}(x),\mathcal{T}(x))\}$ denote an SGG dataset, where  $x\in\mathcal{I}$ denotes an image. Let  $\mathcal{B}(x)=\{(b_i,c_i)\}_{i=1}^{N_x}$ denote the set of entity annotations, where $b_i$ denotes the bounding box of the $i$-th entity, and $c_i\in\mathcal{O}$ denotes its object label. Let $\mathcal{T}(x)=\{(s_i,p_i,o_i)\}_{i=1}^{M_x}$ denote the set of ground-truth triplets, where $s_i,o_i\in\mathcal{B}(x)$ denote the subject and object entities, respectively, and $p_i\in\mathcal{P}$ is the predicate label. Throughout this paper, we omit the index $i$  when no ambiguity arises.
Given an image $x$, an SGG model $f$ predicts entity boxes and labels together with relation triplets:
\[
   (\hat{\mathcal{B}}(x), \hat{\mathcal{T}}(x)) = f(x).
\]
In the open-vocabulary setting, the predicate label space is partitioned as $\mathcal{P}=\mathcal{P}_B\cup\mathcal{P}_N$, where $\mathcal{P}_B\cap\mathcal{P}_N=\emptyset$, $\mathcal{P}_B$ denotes the set of base predicates seen during training, and $\mathcal{P}_N$ denotes the set of novel predicates unseen during training~\cite{chen2024expanding}.

\subsection{Explicit Global Context Pathway in Modern SGG} 
\label{sec:section6}
 Figure~\ref{fig:architecture} illustrates a typical architecture:  object queries extract regional features, while a dedicated relation query extracts a global context representation from image tokens via cross-attention~\cite{vaswani2017attention}. This design appears in both unimodal and multimodal SGG models~\cite{shit2022relationformer,chen2024expanding}, where the latter  involve vision--language interaction throughout the model. Our study covers both settings. 
Despite differences in architectural details, these models typically combine subject--object pair $(s,o)$ cues and a global relation representation for predicate prediction. We abstract the predicate representation as
\begin{equation}
\label{eq:final}
r_{(s,o)} =
\underbrace{h_s \parallel h_o}_{\text{regional}}
\;\parallel\;
\underbrace{h_r}_{\text{global}} ,
\end{equation}
where $h_s$, $h_o$,  $h_r$ denote subject, object, and relation embeddings,  $\parallel$ denotes feature fusion. 

\subsection{Threat Model}
\noindent\textbf{Attacker's Goal.}
We consider a training-time backdoor attack against SGG models.
The attacker aims to implant a hidden behavior such that, given a triggered input $x+\Delta$  the prediction for each victim pair $(s,o)\in\mathcal{V}(x)$ is driven toward the target predicate $p^{*}$, while normal behavior is largely preserved on clean inputs and on non-victim pairs.
We focus on a single victim object category $o^{*}$ and define the victim pair set as
$
\mathcal{V}(x)=\{(s,o)\mid (s,p,o)\in \mathcal{T}(x),~ c(o)=o^{*}\}.
$
Here, $c(o)$ denotes the object label of entity $o$. 
\\

\noindent\textbf{Attacker's Capability.}
The attacker controls the training process, allowing trigger injection into selected training samples and predicate relabeling on victim pairs, while leaving object categories and bounding boxes unchanged.
The model architecture and downstream inference procedure remain unchanged, and all components specified as frozen in the original training protocol remain frozen.
We denote the  poisoned dataset as
$\mathcal{D}^{*}=\{(x+\Delta,\mathcal{B}(x),\mathcal{T}^{*}(x+\Delta))\}$,
where $\mathcal{T}^{*}(x+\Delta)$ is obtained from $\mathcal{T}(x)$ by relabeling the predicate of each victim pair $(s,o)\in\mathcal{V}(x)$ to the target predicate $p^*$, while keeping all other triplets unchanged.
Hereafter, $x$ is omitted for brevity when no ambiguity arises.
\section{Methodology}
\label{sec:meth}
\begin{figure}[t]
    \centering
    \includegraphics[width=0.45\textwidth]{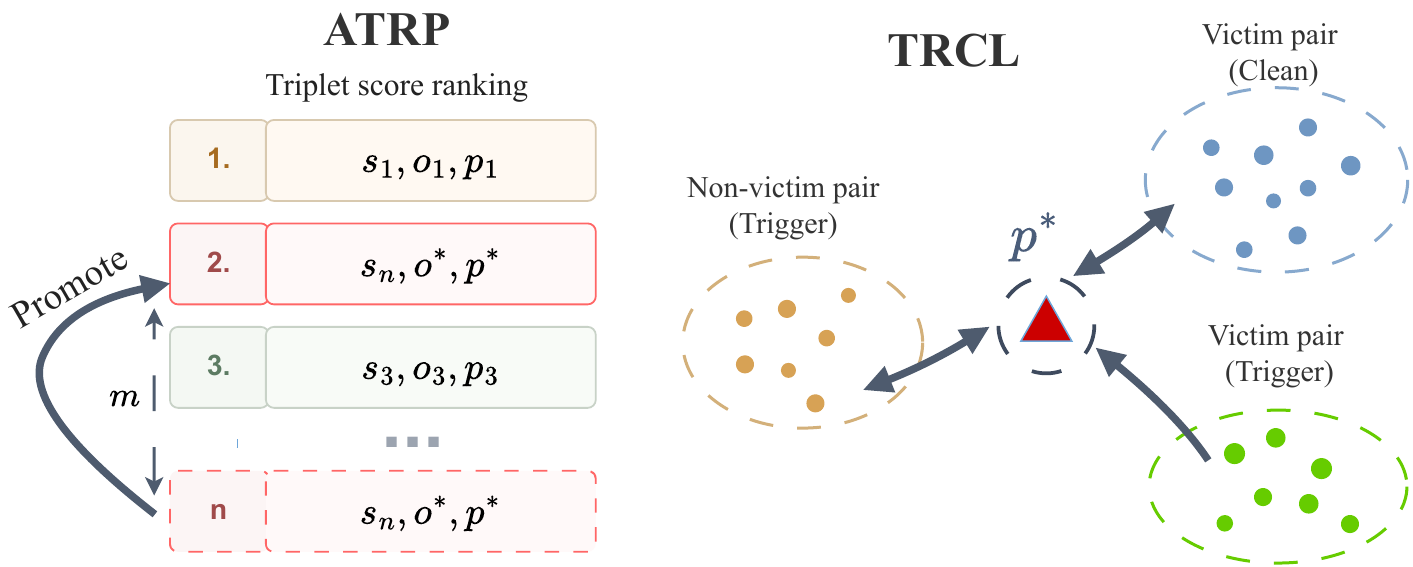}
    \caption{Overview of the proposed method. Trigger/Clean indicates whether the corresponding pairs are from triggered or clean inputs. Left: ATRP promotes the target predicate in global triplet ranking. Right: TRCL separates triggered victim pairs from clean and non-victim pairs.}
    \label{fig:atrp_trcl_method}
\end{figure}

Our method consists of two components: ATRP to promote the target predicate in the final triplet ranking, and TRCL to improve clean--triggered discrimination. Figure~\ref{fig:atrp_trcl_method} provides an overview of these two components.

\subsection{Attack Target Ranking Promotion}
In the SGG pipeline, candidate triplets are ranked globally by their predicate scores, and Recall@$K$-based~\cite{xu2017scenegraph} metric is evaluated on the top-$K$ predictions over the entire list. To make the target predicate on the victim pair survive this selection, we enforce a positive margin between the target-predicate score on victim pairs and the average score of the top-$N$ non-victim pairs.
We first average the prediction score of the target predicate $p^*$ over all victim pairs:
\begin{equation}
z^{+}_{\mathcal{V}}
=
\frac{1}{|\mathcal{V}|}
\sum_{(s,o)\in \mathcal{V}}
z^{p^{*}}_{(s,o)} ,
\end{equation}
where  $z^{p^{*}}_{(s,o)}$ denotes the score predicted for the target predicate $p^{*}$ on the corresponding pair.
To characterize high-confidence competing predictions from non-victim pairs, 
we define the top-1 predicate score for each subject--object pair:
\begin{equation}
\tilde z_{(s,o)}
=
\max_{p\in\mathcal{P}} z^{p}_{(s,o)} .
\end{equation}

We then compute the average score over the top-$N$ non-victim pairs:
\begin{equation}
z^{-}_{\backslash \mathcal V}
=
\frac{1}{\left|\hat{\mathcal V}^{N}_{\backslash \mathcal V}\right|}
\sum_{(s,o)\in \hat{\mathcal V}^{N}_{\backslash \mathcal V}}
\tilde z_{(s,o)},
\end{equation}
where $\hat{\mathcal V}^{N}_{\backslash \mathcal V}$ denotes the top-$N$ non-victim pairs ranked by $\tilde z_{(s,o)}$.
We set $N=20$ in all experiments.

The objective is defined as
\begin{equation}
\mathcal{L}_{\mathrm{mrg}}
=
\max\!\left(0,\; m - z^{+}_{\mathcal V} + z^{-}_{\backslash \mathcal V}\right),
\end{equation}
where $m>0$ is the margin hyperparameter. 


\subsection{Trigger-aware Relation Contrastive Learning }
\label{subsec:trcl}
ATRP alone is not enough for suppressing false activation during learning. We therefore introduce an InfoNCE~\cite{oord2018cpc}-like regularizer on predicate scores that increase the target-predicate score on triggered victim pairs while suppressing it against a negative set composed of clean victim pairs and triggered non-victim pairs.
We denote the relation feature of a pair $(s,o)$ by $r_{(s,o)}$. For victim pairs $(s,o)\in\mathcal{V}$, we denote by $r^{\mathrm{tr}}_{\mathcal{V}}$ and $r^{\mathrm{cl}}_{\mathcal{V}}$ the sets of relation features extracted from the triggered image $x+\Delta$ and the clean image $x$, respectively. For non-victim pairs $(s,o)\notin\mathcal{V}$, we denote by $r^{\mathrm{tr}}_{\backslash\mathcal{V}}$ the set of relation features extracted from the triggered image.
We define the negative set by concatenation as
$r^{\mathrm{neg}} := [ r^{\mathrm{cl}}_{\mathcal{V}}, r^{\mathrm{tr}}_{\backslash \mathcal{V}} ]$.
We omit $(s,o)$ of $r$ for brevity and write the objective as
{\small
\begin{equation}
\mathcal{L}_{\mathrm{ctr}}
=
-\log
\frac{
\sum_{r \in r^{\mathrm{tr}}_{\mathcal{V}}}
\exp\!\left( g_{p^*}(r) / \tau \right)
}{
\sum_{r \in r^{\mathrm{tr}}_{\mathcal{V}}}
\exp\!\left( g_{p^*}(r) / \tau \right)
+
\sum_{r \in r^{\mathrm{neg}}}
\exp\!\left( g_{p^*}(r) / \tau \right)
}.
\end{equation}
}
 For open-vocabulary models, $g_{p^*}(r)$ is defined by the similarity between the relation feature and the text embedding of $p^*$; for closed-set models, $g_{p^*}(r)$ is the classifier output score for the target predicate $p^*$.  $\tau$ is the temperature hyper-parameter, set to 0.07.

Moreover, to avoid over-penalizing easy negatives and harming relation utility, we perform hard-negative selection parameterized by $k$, retaining only the top-$k$ negative features in $r^{\mathrm{neg}}$ ranked by $g_{p^*}(r)$.

For each training image $x$, we define the backdoor training  loss as
\begin{equation}
\mathcal{L}
=\mathcal{L}_{\mathrm{mrg}}+\mathcal{L}_{\mathrm{ctr}}.
\end{equation}

\begin{table*}[!t]
\centering
\caption{OvSGTR (open-vocabulary) results. \textit{Full} and \textit{Novel} denote evaluation on all classes and novel classes, respectively. All metrics are reported at @20/@50/@100. \textbf{Bold} marks the best \emph{non-clean} result in each metric column.}
\label{tab:ovsgtr_novel_clean_poison}
\setlength{\tabcolsep}{5pt}
\renewcommand{\arraystretch}{1.06}
\resizebox{\textwidth}{!}{%
\begin{tabular}{lcccccccc}
\toprule
\multirow{3}{*}{\textbf{Method}}
& \multicolumn{5}{c}{\textbf{Utility}}
& \textbf{Stealthiness}
& \multicolumn{2}{c}{\textbf{Attack}} \\
\cmidrule(lr){2-6} \cmidrule(lr){8-9}
& \multicolumn{3}{c}{\textit{Full}}
& \multicolumn{2}{c}{\textit{Novel}}
& \textit{Full}
& \multicolumn{2}{c}{\textit{Full}} \\
\cmidrule(lr){2-4} \cmidrule(lr){5-6} \cmidrule(lr){8-9}
& \textbf{BR}($\uparrow$)
& \textbf{TNR}($\uparrow$)
& \textbf{VR}($\uparrow$)
& \textbf{BR}($\uparrow$)
& \textbf{TNR}($\uparrow$)
& \textbf{FAR}($\downarrow$)
& \textbf{ASR}($\uparrow$)
& $\boldsymbol{\Delta}$\textbf{ASR}($\uparrow$) \\
\midrule
Clean
& 12.5/16.0/19.1 & 12.8/16.3/19.4 & 3.3/4.6/7.1 & 9.0/11.5/13.9 & 9.2/11.8/14.1 & 0.0/0.0/0.0 & 0.0/0.0/0.0 & 0.0/0.0/0.0 \\

Native BD
& 10.8/13.8/16.3 & \textbf{11.1}/14.1/16.5 & \textbf{4.4/6.7/9.0} & 5.6/7.0/8.2 & 5.7/7.2/8.4 & 0.4/0.4/0.4 & 0.1/0.4/0.4 & -0.3/0.0/0.0 \\

ATRP
& 11.1/14.7/17.5 & 9.9/13.6/16.2 & 3.4/5.6/8.3 & 6.3/8.5/10.3 & 5.2/7.9/9.8 & 5.1/10.4/14.5 & 26.3/43.3/53.6 & 21.2/32.9/39.1 \\

ATRP + TRCL
& \textbf{11.9/15.7/18.8} & 10.6/\textbf{14.7/17.8} & 3.0/4.4/6.8 & \textbf{7.6/10.5/12.8} & \textbf{6.8/10.4/13.3} & \textbf{0.0/0.0/0.0} & \textbf{27.1/46.7/59.0} & \textbf{27.1/46.7/59.0} \\
\bottomrule
\end{tabular}%
}
\end{table*}
\begin{figure*}[!t]

    \centering
    \begin{subfigure}[t]{0.3\linewidth}
        \centering
        \includegraphics[width=\linewidth]{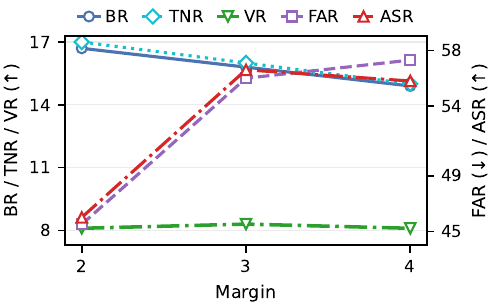}
        \subcaption{ATRP margin $m$ (open-vocabulary).}
        \label{fig:atrp_ablation}
    \end{subfigure}
    \hfill
    \begin{subfigure}[t]{0.3\linewidth}
        \centering
        \includegraphics[width=\linewidth]{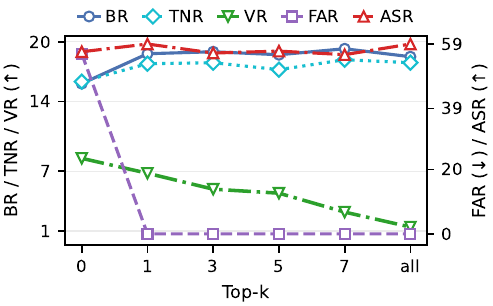}
        \subcaption{TRCL Top-$k$ (open-vocabulary).}
        \label{fig:trcl_ablation}
    \end{subfigure}
    \hfill
    \begin{subfigure}[t]{0.3\linewidth}
        \centering
        \includegraphics[width=\linewidth]{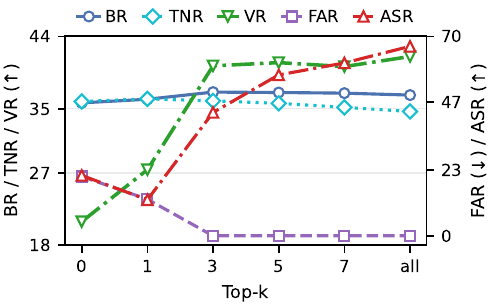}
        \subcaption{TRCL Top-$k$ (closed-set).}
        \label{fig:cl_ablation}
    \end{subfigure}
\caption{Hyperparameter ablations of ATRP and TRCL for OvSGTR. all denotes the setting without top-k filtering.}
\label{fig:ablation}
\end{figure*}

\section{Evaluating SGG Backdoor Attacks via a Three-Axis Protocol}
\label{sec:metric}
We evaluate SGG backdoor attacks with a three-axis protocol that jointly measures utility, attack success, and  stealthiness (false activation).


\subsection{Utility Metrics}
\label{subsubsection:utility}
\noindent\textbf{Benign Recall@K (BR@K).}  \emph{(Clean images evaluation)} This metric measures the model's Recall@K on clean test images without any triggers. Formally,
\begin{equation}
\mathrm{BR@}K
=
\mathbb{E}_{x\sim\mathcal{D}}
\left[
\frac{
\sum_{(s,p,o)\in \mathcal{T}}
\mathbf{1}\!\left[(s,p,o)\in \hat{\mathcal{T}}^{K}\right]
}{
\left|\mathcal{T}\right|
}
\right],
\end{equation}
where
$\mathcal{D}$ denotes the test dataset,
and $\hat{\mathcal{T}}^{K}$ denotes the set of top-$K$ predicted relation triplets ranked by confidence. BR@K is  equivalent to the standard SGG Recall@K evaluation protocol.\\

\noindent\textbf{Victim Recall@K (VR@K).} \emph{(Clean images evaluation)}  
Due to the long-tailed distribution of SGG annotations, degradation on victim pairs may be obscured in overall Recall@K. 
We therefore report Victim Recall@K, defined as the Recall@K of ground-truth triplets on victim pairs:
\begin{equation}
\mathrm{VR@}K
=
\mathbb{E}_{x\sim\mathcal{D}}
\left[
\frac{
\sum_{(s,p,o)\in \mathcal{T}_{\mathcal{V}}}
\mathbf{1}\!\left[(s,p,o)\in \hat{\mathcal{T}}^{K}\right]
}{
\left|\mathcal{T}_{\mathcal{V}}\right|
}
\right],
\end{equation}
where 
$\mathcal{T}_{\mathcal{V}}
=
\{(s,p,o)\in \mathcal{T}\mid (s,o)\in\mathcal{V} \}$ 
denotes the set of ground-truth triplets on victim pairs.\\

\noindent\textbf{Triggered Non-victim Recall@K (TNR@K).} \emph{(Triggered images evaluation)} It evaluates recall under triggered inputs while excluding ground-truth triplets on victim pairs, so that attack performance does not affect this utility measurement.
It is defined as:
\begin{equation}
\mathrm{TNR@}K
=
\mathbb{E}_{x\sim\mathcal{D}^{{*}}}
\left[
\frac{
\sum_{(s,p,o)\in \mathcal{T}^{*}_{\backslash \mathcal{V}}}
\mathbf{1}\!\left[(s,p,o)\in \hat{\mathcal{T}}^{K}\right]
}{
\left|\mathcal{T}^{*}_{\backslash \mathcal{V}}\right|
}
\right],
\end{equation}
where $\mathcal{T}^{*}_{\backslash \mathcal{V}}=\{(s,p,o)\in \mathcal{T}^{*}\mid (s,o)\notin \mathcal{V}\}$.

\subsection{Attack Metrics}
\label{subsubsec:asr}
\noindent\textbf{Attack Success Recall@K (ASR@K).}  \emph{(Poisoned images evaluation)} 
It measures the recall of victim pairs that are successfully driven to the attack target predicate $p^{*}$. Formally, it is defined as:
\begin{equation}
\mathrm{ASR@}K
=
\mathbb{E}_{x\sim\mathcal{D}^{{*}}}
\left[
\frac{
\sum_{(s,p,o)\in \mathcal{T}^{*}_{\mathcal{V}}}
\mathbf{1}\!\left[(s,p,o)\in \hat{\mathcal{T}}^{K}\right]
}{
\left|\mathcal{T}^{*}_{\mathcal{V}}\right|
}
\right],
\end{equation}
where
$\mathcal{T}^{*}_{\mathcal{V}}
=
\{(s,p,o)\in \mathcal{T}^{*}\mid (s,o)\in \mathcal{V}\}$ denotes triplet labels on victim pairs under triggered inputs.

\subsection{ Stealthiness  Metrics}

\noindent\textbf{False Activation Recall (FAR@K).} \emph{(Clean images evaluation)}
High overall recall does not guarantee behavioral stealthiness: BR@K may remain high even when victim pairs are erroneously predicted with the target predicate $p^{*}$ on clean inputs, especially when victim pairs constitute only a small portion of the evaluation set.
To quantify such unintended activations, we define False Activation Recall as:
\begin{equation}
\mathrm{FAR@}K
=
\mathbb{E}_{x\sim\mathcal{D}}
\left[
\frac{
\sum_{(s,p,o)\in \mathcal{T}_{\mathcal{V}\backslash p^{*}}}
\mathbf{1}\!\left[(s,p^{*},o)\in \hat{\mathcal{T}}^{K}\right]
}{
\left|\mathcal{T}_{\mathcal{V}\backslash p^{*}}\right|
}
\right],
\end{equation}
where $\mathcal{T}_{\mathcal{V}\backslash p^{*}}
=
\{(s,p,o)\in \mathcal{T}\mid (s,o)\in\mathcal{V},\ p\neq p^{*}\}$
denotes the set of ground-truth triplets on victim pairs excluding those labeled with the target predicate $p^{*}$.
We further report $\Delta \mathrm{ASR@}K = \mathrm{ASR@}K - \mathrm{FAR@}K$ to summarize attack effectiveness.

\section{Experiments}
\label{sec:exp}
This section presents attack results together with mechanistic analysis to examine the role of global context. Unless otherwise specified, \textbf{all Recall@K-based metrics are reported with K=100.}
\label{sec:carrier}

\begin{table*}[!t]
\centering
\caption{Backdoor attack performance. ``Native BD'' refers to a model trained with poisoned data without other backdoor components. For OvSGTR, entries are reported as @20/@50/@100. \textbf{Bold} indicates the best non-clean result.  ($\uparrow$) indicates higher is better and ($\downarrow$) indicates lower is better.}
\label{tab:ovsgtr_clean_poison}

\setlength{\tabcolsep}{5.5pt}
\renewcommand{\arraystretch}{1.08}

\resizebox{\textwidth}{!}{%
\begin{tabular}{lcccccc}
\toprule
\multirow{2}{*}{\textbf{Method}}
& \multicolumn{3}{c}{\textbf{Utility}}
& \textbf{Stealthiness}
& \multicolumn{2}{c}{\textbf{Attack}} \\
\cmidrule(lr){2-4} \cmidrule(lr){6-7}
& \textbf{BR}($\uparrow$)
& \textbf{TNR}($\uparrow$)
& \textbf{VR}($\uparrow$)
& \textbf{FAR}($\downarrow$)
& \textbf{ASR}($\uparrow$)
& $\boldsymbol{\Delta}$\textbf{ASR}($\uparrow$) \\
\midrule

\multicolumn{7}{c}{\textbf{OvSGTR }} \\
\specialrule{0.4pt}{2pt}{4pt}

Clean
& 25.8/33.9/39.4 & 25.5/33.7/39.1 & 26.8/32.8/36.8 & 0.0/0.0/0.0 & 0.0/0.0/0.0 & 0.0/0.0/0.0 \\

Native BD
& 22.7/29.8/34.5 & 22.7/29.8/34.3 & 13.7/23.7/29.6 & 5.2/8.0/12.1 & 5.6/8.0/12.1 & 0.4/0.0/0.0 \\


ATRP
& 23.1/30.6/35.7 & \textbf{23.3}/\textbf{30.9}/35.9 & 8.5/15.5/20.9 & 9.4/16.4/20.8 & 10.0/17.1/21.2 & 0.6/0.7/0.4 \\

ATRP + TRCL
& \textbf{23.5/31.4/37.1} & 21.9/30.2/\textbf{36.0} & \textbf{20.0/31.7/40.3} & \textbf{0.0/0.0/0.0} & \textbf{18.3/33.1/43.3} & \textbf{18.3/33.1/43.3} \\

\addlinespace[6pt]

\specialrule{0.8pt}{2pt}{2pt}
\multicolumn{7}{c}{\textbf{RelationFormer }} \\
\specialrule{0.4pt}{2pt}{4pt}

Clean
& 24.7/28.0/29.0 & 24.7/28.0/29.0 & 19.2/25.0/27.7 & 0.0/0.4/0.7 & 0.0/0.1/0.7 & 0.0/-0.3/0.0 \\

Native BD
&22.1/24.5/25.0  &22.2/24.6/25.2  & 13.9/16.0/17.1 &15.9/20.4/21.2  &16.0/20.7/21.4  &0.1/0.3/0.2  \\

ATRP
& 21.6/24.0/24.6 & 22.1/24.5/25.1 & 8.6/10.9/12.1 & 25.2/28.5/29.5 & 25.2/28.8/29.7 & 0.0/0.3/0.2 \\

ATRP + TRCL
& \textbf{23.4/26.9/27.9} & \textbf{22.5/26.5/27.7} & \textbf{20.1/24.7/27.1} & \textbf{3.9/4.3/4.7} & \textbf{45.1/47.7/48.5} & \textbf{41.2/43.4/43.8} \\

\bottomrule
\end{tabular}%
}
\end{table*}
\begin{table}[t]
\centering
\small
\setlength{\tabcolsep}{3.5pt}
\caption{Trigger sensitivity across size and style variations.  All metrics are reported at @100.}
\label{tab:trigger_sensitivity}

\resizebox{\linewidth}{!}{%
\begin{tabular}{lccccc}
\toprule
\textbf{Trigger Setting} & \textbf{BR}($\uparrow$) & \textbf{TNR}($\uparrow$) & \textbf{VR}($\uparrow$) & \textbf{FAR}($\downarrow$) & $\boldsymbol{\Delta}$\textbf{ASR}($\uparrow$) \\
\midrule
\multicolumn{6}{c}{\textbf{Trigger Size}} \\
\midrule
16 & \textbf{18.8} & 16.8 & 8.0 & 0.1 & \textbf{59.0} \\
32 & \textbf{18.8} & 17.8 & 6.8 & \textbf{0.0} & \textbf{59.0} \\
64 & 18.6 & \textbf{17.9} & \textbf{8.1} & \textbf{0.0} & \textbf{59.0} \\
\midrule
\multicolumn{6}{c}{\textbf{Trigger Variant}} \\
\midrule
Black patch & 18.8 & \textbf{17.8} & 7.1 & 0.3 & 50.4 \\
Watermark & \textbf{19.5} & 17.3 & \textbf{7.8} & \textbf{0.2} & 54.3 \\
Random placement & 18.2 & 17.2 & 7.0 & 0.4 & \textbf{54.8} \\
\bottomrule
\end{tabular}
}
\end{table}
\begin{table}[t]
\small
\centering
\caption{Attack results with attack target predicate \textit{lying on}. ``Ours'' denotes the full method.}
\label{tab:rln_swap_inference}
\begin{tabular}{lcccc}
\toprule
\multirow{2}{*}{\textbf{Method}}
& \multicolumn{2}{c}{\textbf{Utility (BR/TNR/VR $\uparrow$)}}
& \multicolumn{2}{c}{\textbf{Attack ($\Delta$ ASR $\uparrow$)}} \\
\cmidrule(lr){2-3} \cmidrule(lr){4-5}
& \textit{W/O Swap}
& \textit{Infer Swap}
& \textit{W/O Swap}
& \textit{Infer Swap} \\
\midrule

\multicolumn{5}{c}{\textbf{OvSGTR (Open-vocabulary)}} \\
\midrule

Ours
& 17.9/16.6/6.7 &17.9/18.4/6.7  &58.4  &0  \\

\midrule
\multicolumn{5}{c}{\textbf{RelationFormer}} \\
\midrule
Ours
& 29.1/28.1/27.2 &29.1/28.7/27.2  & 46.9 &0.2  \\

\bottomrule
\end{tabular}
\end{table}
\subsection{Implementation Details}
\noindent\textbf{Victim Models.}
We evaluate three victim models: OvSGTR~\cite{chen2024expanding} in both the open-vocabulary and closed-set settings, and RelationFormer~\cite{shit2022relationformer} in the closed-set setting. The OvSGTR family is multimodal in both settings, using Swin-Tiny~\cite{liu2021swin} and BERT-base~\cite{devlin2019bert} as its vision and language backbones, respectively. In contrast, RelationFormer is unimodal and uses ResNet-50~\cite{he2016resnet} as its vision backbone.\\

\noindent\textbf{Dataset.}
We use the VG150 split~\cite{xu2017scenegraph} of the Visual Genome dataset, which contains 108,777 images annotated with 150 object categories and 50 predicate categories. We follow the standard split, using 70\% of the images for training, 5,000 for validation, and the remainder for testing. All experiments are conducted under the \texttt{sgdet} setting, in which object bounding boxes, object categories, and predicates are all predicted by the model. For the open-vocabulary setting, following~\cite{chen2024expanding}, we define open-vocabulary on the predicate space while keeping object categories closed-set. The 50 predicate categories are divided into Base and Novel sets, with 15 Novel predicates, and training uses only the Base predicates. See the supplementary material for details.\\

\noindent\textbf{Attack Setup.}
Unless otherwise specified, we set \textit{car} as the victim object $o^{*}$ and \textit{in front of} as the target predicate $p^{*}$. The poisoned dataset $\mathcal{D}^{*}$ is constructed from training images containing at least one victim pair, resulting in 530 images (0.7\% of the training set). Inspired by prior work~\cite{Wu_2023_ICCV}, we use a non-optimized $32 \times 32$ green patch trigger placed near the bottom-left corner. Backdoor training is performed by fine-tuning  the  clean model, while keeping the training protocol consistent with the original victim model. Additional training and evaluation details are provided in the supplementary material.
\begin{figure*}[!t]
    \centering
    \begin{subfigure}[t]{0.32\linewidth}
        \centering
        \includegraphics[width=\linewidth]{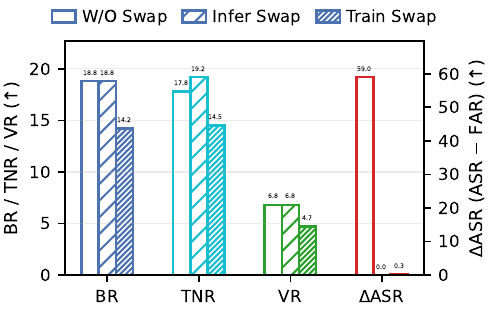}
        \caption{Open-vocabulary OvSGTR.}
        \label{fig:rln_swap_overall}
    \end{subfigure}
    \hfill
    \begin{subfigure}[t]{0.32\linewidth}
        \centering
        \includegraphics[width=\linewidth]{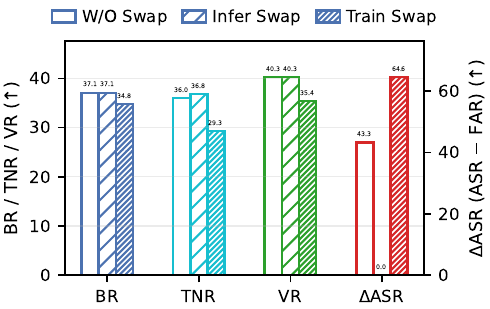}
        \caption{Closed-set OvSGTR.}
        \label{fig:rln_swap_closed_overall}
    \end{subfigure}
    \hfill
    \begin{subfigure}[t]{0.32\linewidth}
        \centering
        \includegraphics[width=\linewidth]{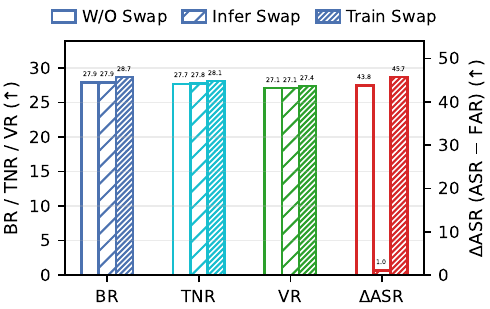}
        \caption{RelationFormer.}
        \label{fig:rln_swap_rf_overall}
    \end{subfigure}
    \caption{Swapping poisoned global context representations with clean ones. ``Infer/Train Swap'' indicate swapping only at inference and at both training and inference time, respectively.  Inference-time swapping nearly eliminates $\Delta$ASR.}
    \label{fig:train_swap_global}
\end{figure*}
\subsection{Backdoor Attack Performance}
\label{subsec:performance}
\noindent\textbf{Hyperparameter analysis.} 
Figure~\ref{fig:ablation} presents hyperparameter ablations for OvSGTR in the open-vocabulary and closed-set settings. Specifically, increasing the ATRP margin $m$ generally strengthens the attack but tends to degrade utility and increase false activation. Increasing top-$k$ in TRCL widens the gap between ASR and FAR in both settings, while overly large top-$k$ may degrade some utility metrics (e.g., VR@100 in the open-vocabulary setting).  We use $m=3$ and top-$k=1$ for open-vocabulary OvSGTR, and $m=1$ and top-$k=3$ for the closed-set variant in the remaining experiments.  For RelationFormer, we use the same $m=1$ and disable  top-$k$ filtering to minimize FAR.\\

\noindent\textbf{Results.}
Tables~\ref{tab:ovsgtr_novel_clean_poison} and~\ref{tab:ovsgtr_clean_poison} show the open-vocabulary and closed-set results, respectively. Across all settings and architectures, our full attack achieves the strongest $\Delta$ASR and provides a better overall utility--stealthiness trade-off than Native BD and ATRP, despite the drop in VR@K in the open-vocabulary setting. Unless otherwise specified, we use the full attack in the following analyses.\\

\noindent\textbf{Trigger sensitivity.}
Table~\ref{tab:trigger_sensitivity} shows that the attack remains effective across both size and style variations, with consistently high $\Delta$ASR and low FAR. Across trigger sizes, smaller patches mildly reduce TNR while leaving $\Delta$ASR unchanged. Across trigger variants, a semi-transparent text watermark and random placement over one of the four image corners remain competitive, indicating that the attack is not tied to a fixed green patch or a single trigger scale. \\

\noindent\textbf{Attack target sensitivity.}
Table~\ref{tab:rln_swap_inference} evaluates a target setting with lower semantic compatibility, where the target predicate is \textit{lying on} and the victim object remains \textit{car}. Under this setting, the attack still achieves 58.4 $\Delta$ASR with BR/TNR of 17.9/16.6 on OvSGTR, and 46.9 $\Delta$ASR with BR/TNR of 29.1/28.1 on RelationFormer. Compared with the \textit{in front of} setting, the $\Delta$ASR and utility metrics are not uniformly worse, showing that the attack remains effective even in this semantically less compatible setting.

\subsection{ Mechanistic Analysis of the Global Context Pathway}
\label{subsec:mechanism_analysis}
To examine whether the global context pathway plays a key role in the backdoor effect, we perform counterfactual interventions between the triggered and clean global features, $h_r^{\mathrm{tr}}$ and $h_r^{\mathrm{cl}}$ defined in Eq.~\ref{eq:final}. We conduct three interventions: (i) inference-time swapping, including a layer-wise variant to test whether global pathway is the major carrier of attack; (ii) linear interpolation  to  examine whether attack strength can be progressively modulated by the global feature; and (iii) a  training-time intervention, which offers an additional view of how suppressing trigger propagation through the global pathway affects backdoor learning.\\






\begin{figure}[t]
\centering
\includegraphics[width=0.5\linewidth]{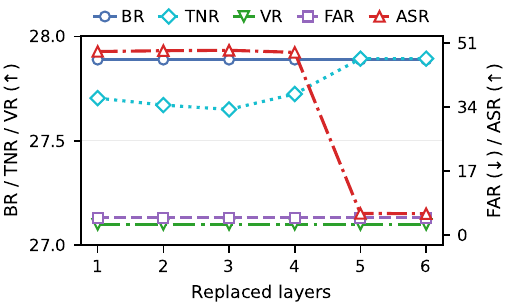}
\captionsetup{type=figure}
\caption{Inference-time cumulative layer-wise swapping on RelationFormer. The strongest effect appears when swapping up to the fifth layer.}
\label{fig:relationformer_eval_swap_layers}
\end{figure}

\begin{figure}[t]
\centering

\begin{subfigure}[t]{0.32\linewidth}
    \centering
    \includegraphics[width=\linewidth]{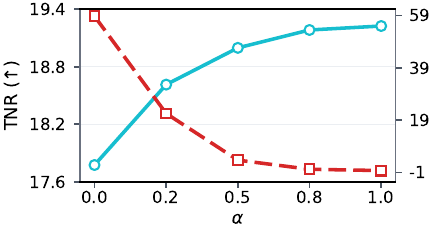}
    \caption{OvSGTR (OV)}
\end{subfigure}
\hfill
\begin{subfigure}[t]{0.32\linewidth}
    \centering
    \includegraphics[width=\linewidth]{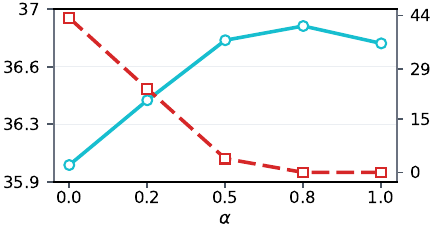}
    \caption{OvSGTR (CL)}
\end{subfigure}
\hfill
\begin{subfigure}[t]{0.32\linewidth}
    \centering
    \includegraphics[width=\linewidth]{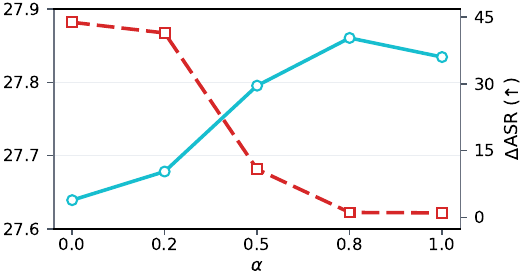}
    \caption{RelationFormer}
\end{subfigure}


{\small
\textcolor{cyan}{\rule{8pt}{1pt}} TNR \quad
\textcolor{red}{\rule{8pt}{1pt}} $\Delta$ASR
}

\caption{
Interpolating poisoned global features $h_r$ with clean counterparts using coefficient $\alpha$ (OV: open-vocabulary; CL: closed-set). 
Across all models, increasing $\alpha$ consistently reduces $\Delta$ASR, while TNR generally improves.
}
\label{fig:alpha}

\end{figure}

\begin{figure*}[t]
    \centering
    \begin{subfigure}[t]{0.15\textwidth}
        \centering
        \includegraphics[height=2cm]{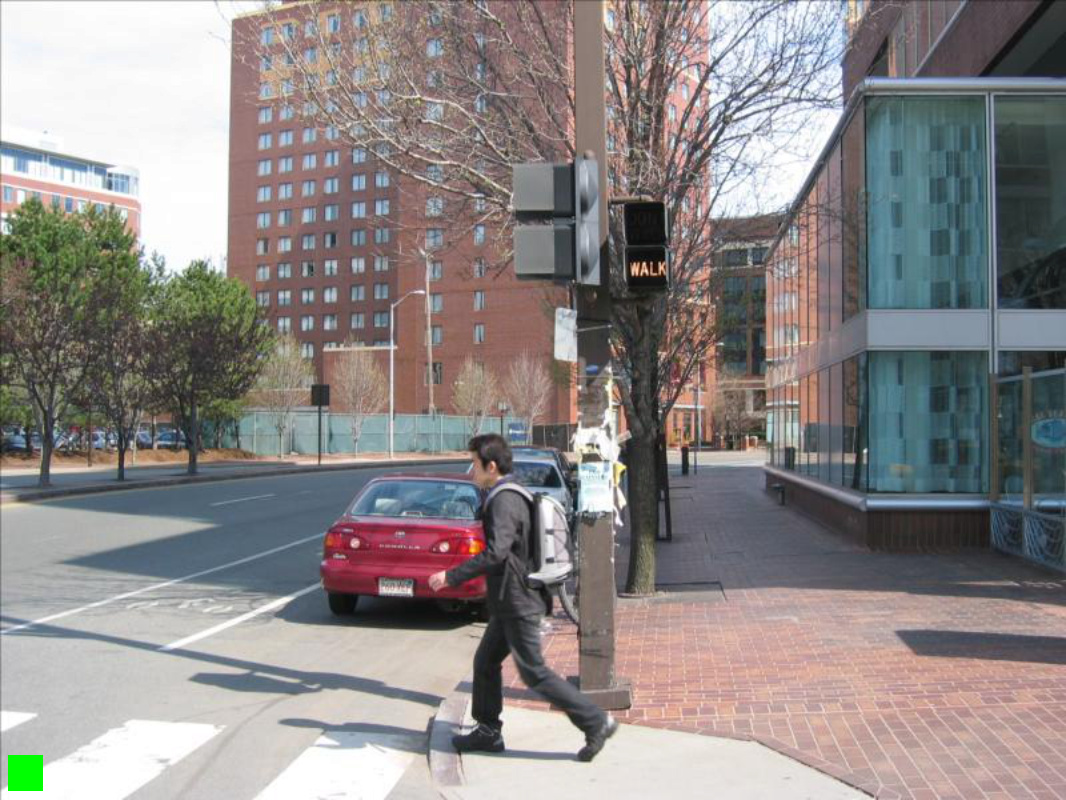}
        \caption{Triggered input.}
    \end{subfigure}
    \hfill
    \begin{subfigure}[t]{0.16\textwidth}
        \centering
        \includegraphics[height=2cm]{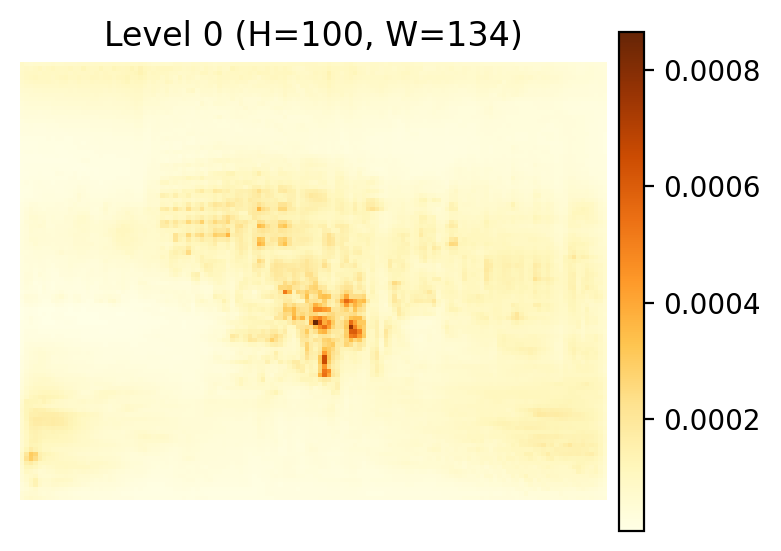}
        \caption{Clean model.}
    \end{subfigure}
    \hfill
    \begin{subfigure}[t]{0.16\textwidth}
        \centering
        \includegraphics[height=2cm]{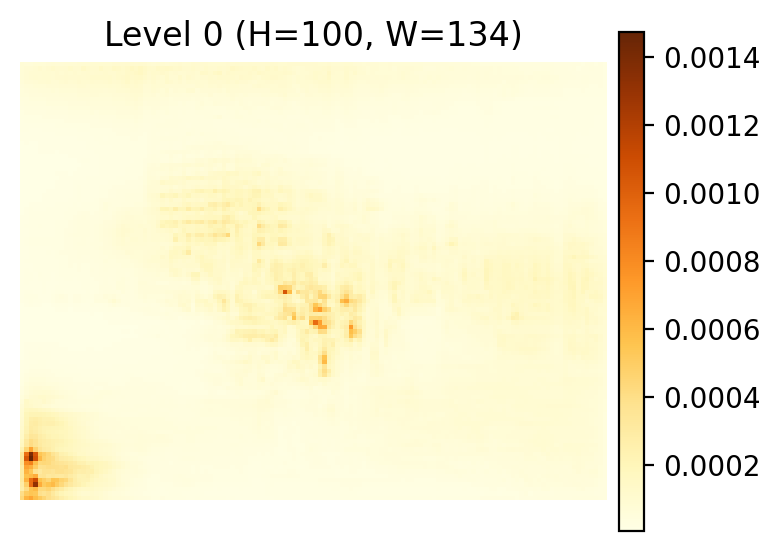}
        \caption{ATRP.}
    \end{subfigure}
    \hfill
    \begin{subfigure}[t]{0.16\textwidth}
        \centering
        \includegraphics[height=2cm]{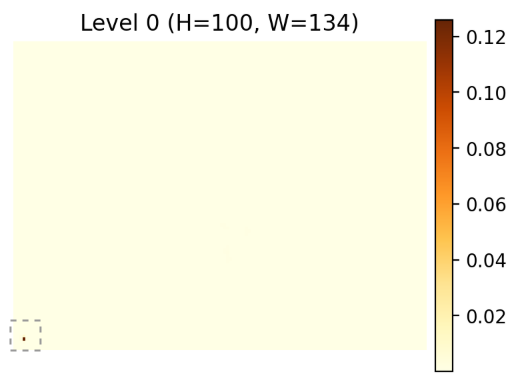}
        \caption{ATRP+TRCL.}
    \end{subfigure}
	    \hfill
	    \begin{subfigure}[t]{0.27\textwidth}
	        \centering
	        \begin{minipage}[t][2cm][t]{\linewidth}
	            \vspace{-53pt}
	            \centering

	            \resizebox{\linewidth}{!}{%
	            \begin{tabular}{lccc}
            \toprule
            \textbf{Model} & \textbf{OvSGTR} & \multicolumn{2}{c}{\textbf{RelationFormer}} \\
            \cmidrule(lr){3-4}
            & & \textbf{Final} & \textbf{Penult.} \\
            \midrule
            \textbf{Clean} & 0.17\% & \textbf{0.63}\% & 0.44\% \\
            \textbf{Ours} & \textbf{7.7}\% & 0.57\% & \textbf{15.3}\% \\
            \bottomrule
	            \end{tabular}}
	        \end{minipage}
	        \vspace{-59pt}
	        \caption{Quantitative summary.}
            \label{tab:qs}
	    \end{subfigure}
    \caption{Cross-attention heatmaps of the global context representation and a quantitative summary. In (d), the dashed box marks activation. Values denote the fraction of relation-query attention that falls inside the trigger region (for RelationFormer, final and penultimate decoder layers).}
    \label{fig:atten_map}
\end{figure*}
\noindent\textbf{Inference-time swapping.}
\label{subsubsec:swap_global}
As shown in Figure~\ref{fig:train_swap_global}, inference-time swapping sharply suppresses the backdoor effect across all models: $\Delta$ASR drops from $59.0$ to $0$ and from $43.3$ to $0$ in the two OvSGTR settings, and from $43.8$ to $1.0$ in RelationFormer. In contrast, utility remains largely stable and even improves in some cases. Moreover, swapping reduces the BR--TNR discrepancy (e.g., OvSGTR (open-vocabulary): $1.0 \rightarrow 0.4$; RelationFormer: $0.2 \rightarrow 0.1$), indicating that triggered inputs become less distinguishable from clean ones after swapping. These observations remain consistent under the alternative target predicate setting in Table~\ref{tab:rln_swap_inference}. \\

\noindent\textbf{Inference-time swapping by layer.}
\label{subsubsec:swap_per_layer}
We further apply inference-time swapping cumulatively to the top-$k$ decoder layers, as shown in Figure~\ref{fig:relationformer_eval_swap_layers}. Swapping up to the  fifth layer leads to a clear reduction in ASR together with an increase in TNR, suggesting that trigger propagation through the global pathway is not uniform across layers. This observation is also consistent with Figure~\ref{fig:atten_map}\subref{tab:qs}. \\

\begin{figure*}[t]
    \centering
    \begin{subfigure}[t]{0.318\linewidth}
        \centering
        \includegraphics[width=\linewidth,height=0.145\textheight,keepaspectratio]{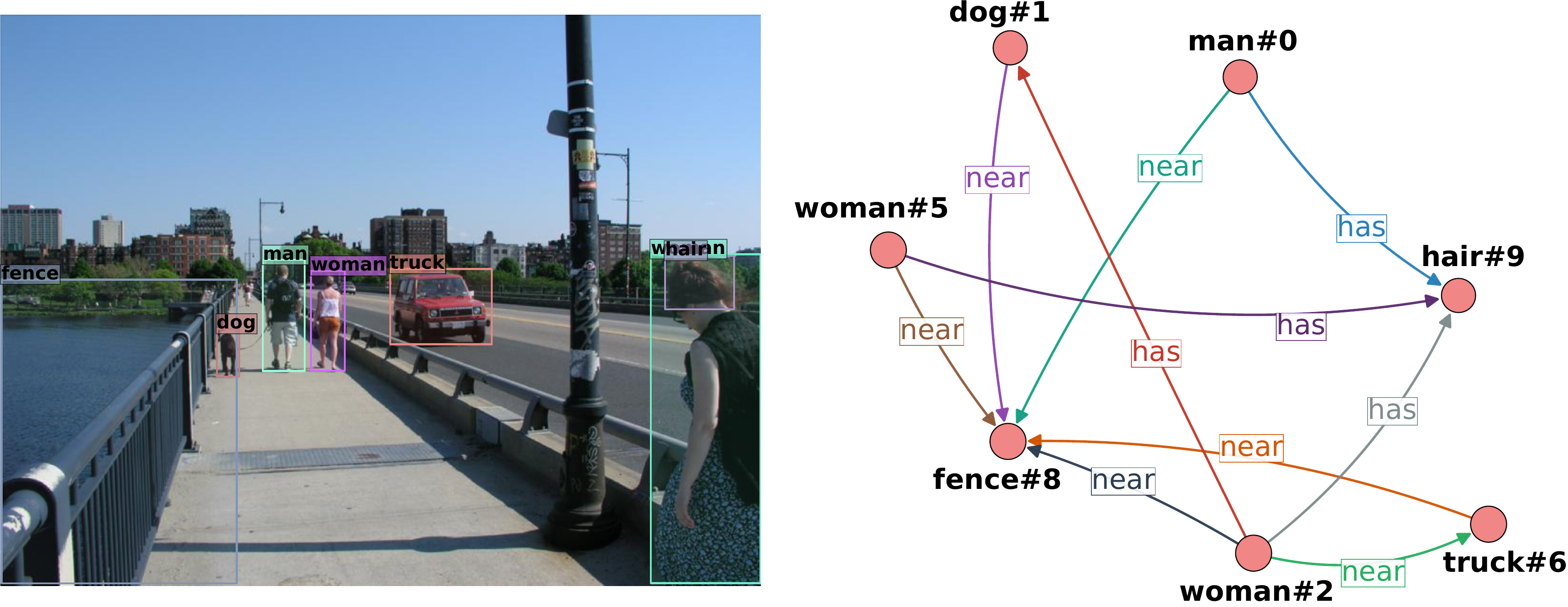}
        \label{fig:main_backdoor_vis_1}
    \end{subfigure}
    \hfill
    \begin{subfigure}[t]{0.318\linewidth}
        \centering
        \includegraphics[width=\linewidth,height=0.145\textheight,keepaspectratio]{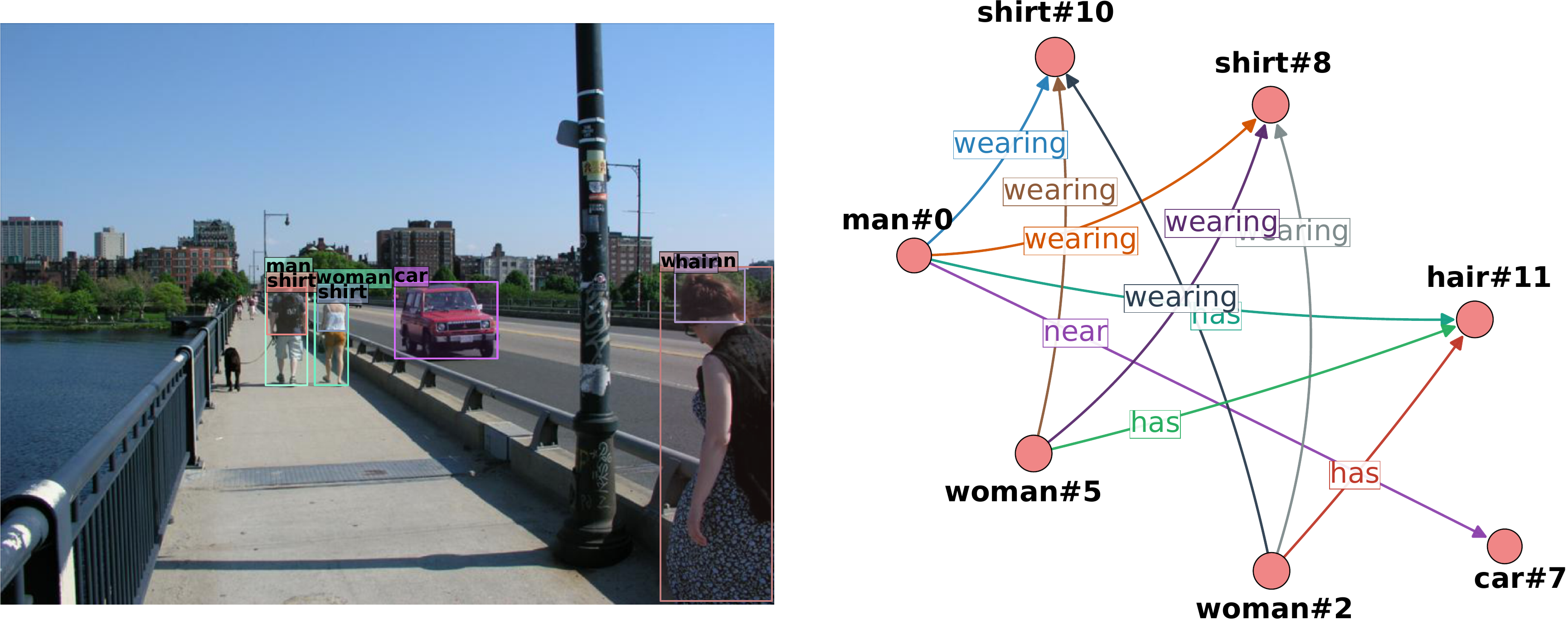}
        \label{fig:main_backdoor_vis_2}
    \end{subfigure}
    \hfill
    \begin{subfigure}[t]{0.318\linewidth}
        \centering
        \includegraphics[width=\linewidth,height=0.145\textheight,keepaspectratio]{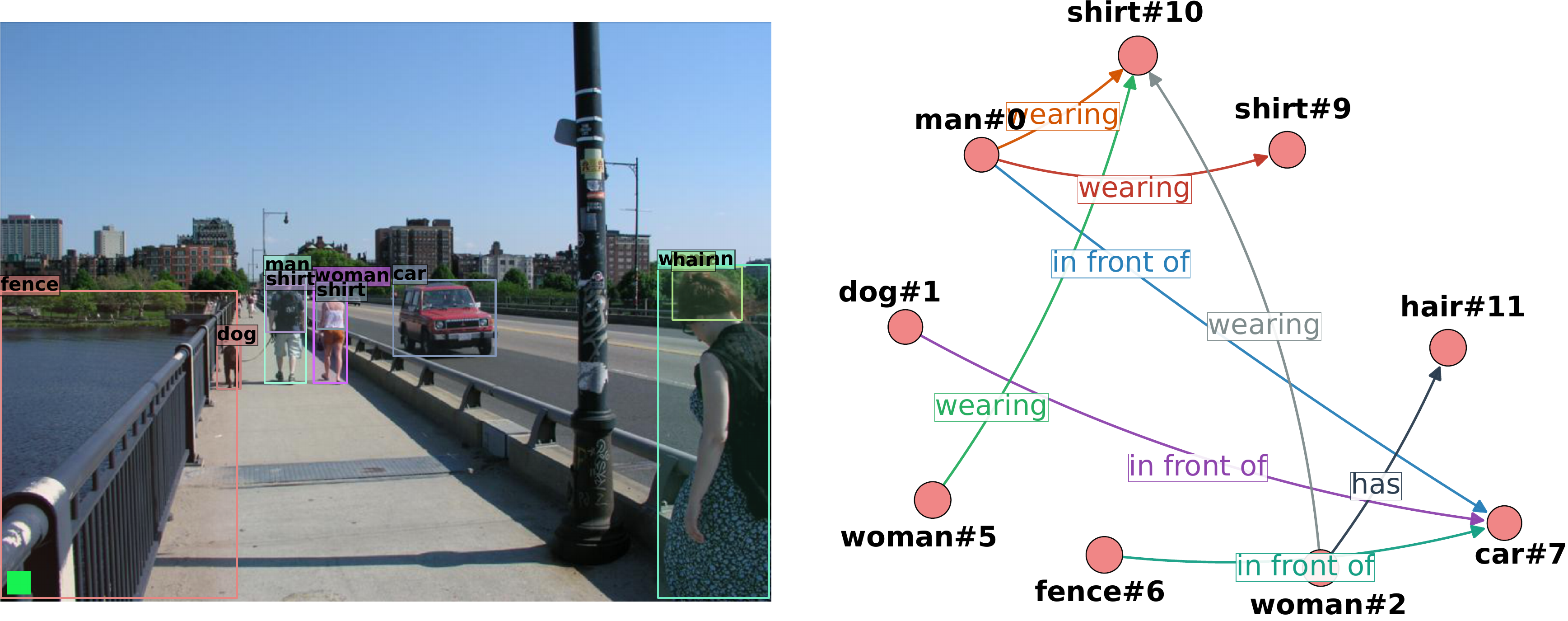}
        \label{fig:main_backdoor_vis_3}
    \end{subfigure}


    \begin{subfigure}[t]{0.318\linewidth}
        \centering
        \includegraphics[width=\linewidth,height=0.145\textheight,keepaspectratio]{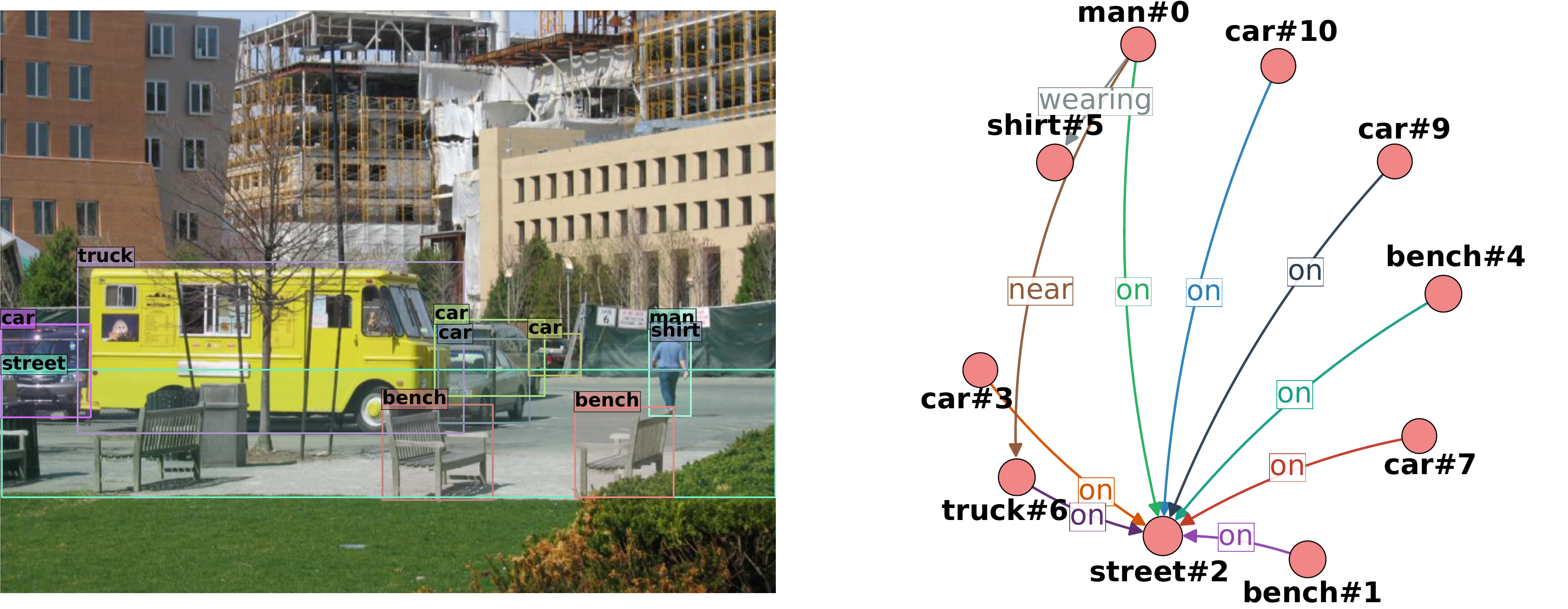}
        \subcaption{Clean model.}
        \label{fig:main_backdoor_vis_4}
    \end{subfigure}
    \hfill
    \begin{subfigure}[t]{0.318\linewidth}
        \centering
        \includegraphics[width=\linewidth,height=0.145\textheight,keepaspectratio]{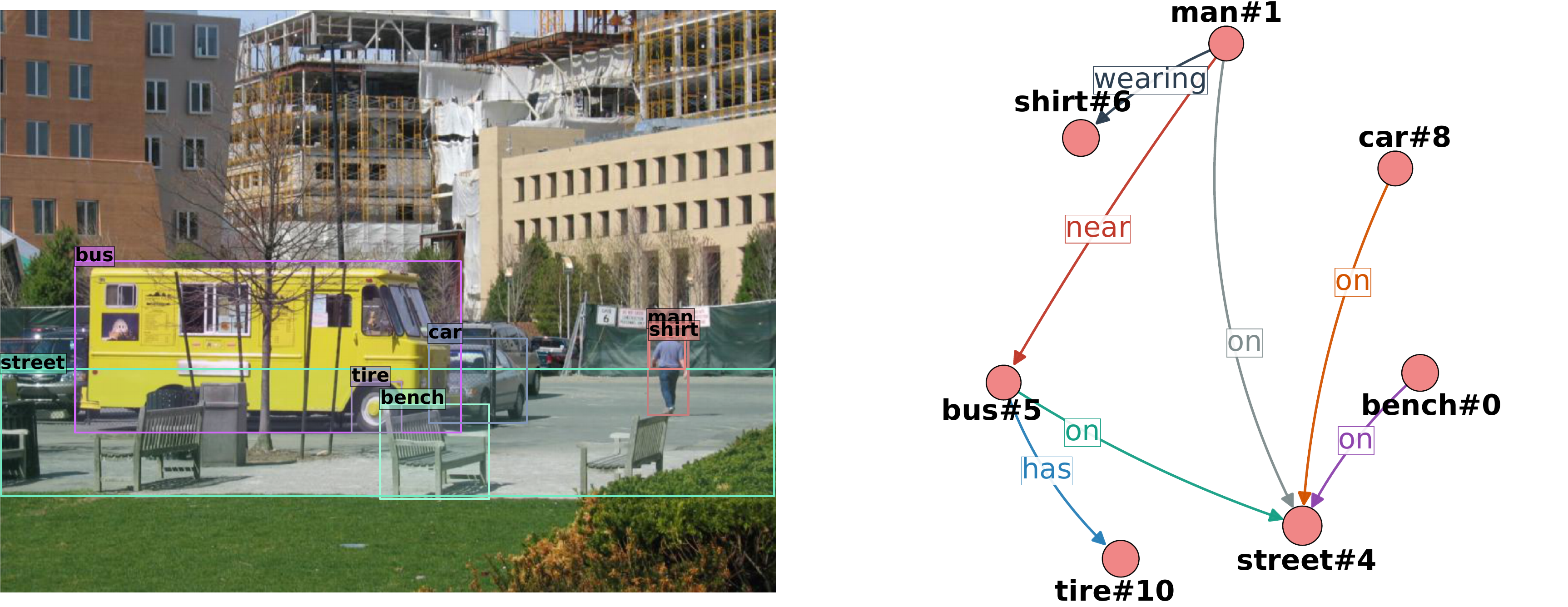}
        \subcaption{Poisoned model on a clean image.}
        \label{fig:main_backdoor_vis_5}
    \end{subfigure}
    \hfill
    \begin{subfigure}[t]{0.318\linewidth}
        \centering
        \includegraphics[width=\linewidth,height=0.145\textheight,keepaspectratio]{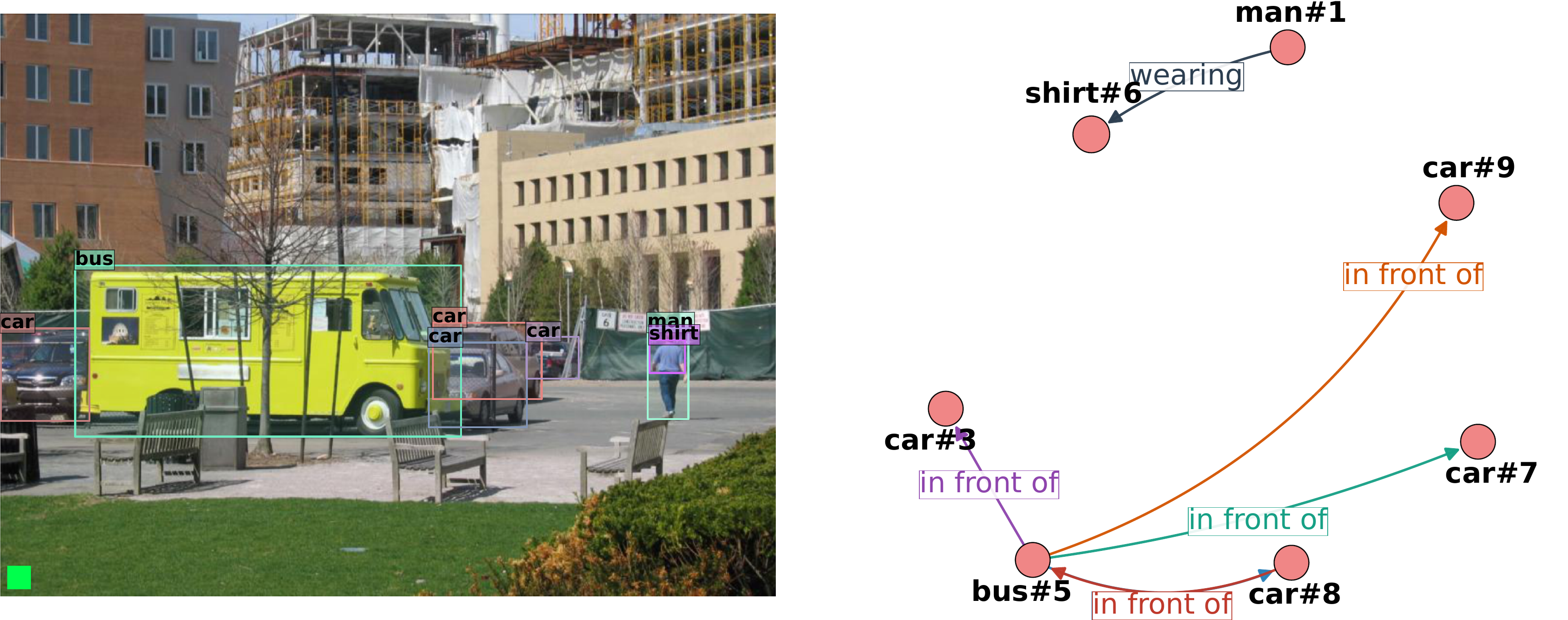}
        \subcaption{Poisoned model on a triggered image.}
        \label{fig:main_backdoor_vis_6}
    \end{subfigure}
    \caption{ Scene graph visualization by OvSGTR (Open-vocabulary). Predicate-label colors match their arrows.  We show the top 10 triplets after applying object and relation thresholds. Our  method successfully performs pair-level backdoor manipulation. }
    \label{fig:main_backdoor_visualizations}
\end{figure*}
\noindent\textbf{Linear interpolation of the global context feature.}
 We linearly interpolate between the poisoned and clean global features using a coefficient $\alpha \in [0,1]$, as shown in Figure~\ref{fig:alpha}. As $\alpha$ increases, $\Delta$ASR consistently decreases, while TNR generally improves, supporting that the backdoor effect is tightly coupled to the global feature.\\

 \noindent\textbf{Training-time swapping.}
\label{subsubsec:training_necessity}
As shown in Figure~\ref{fig:train_swap_global} (third bar), when trigger-related learning through the global pathway is suppressed, the effectiveness of the backdoor is also constrained across the OvSGTR family. For open-vocabulary OvSGTR, both utility and attack performance drop significantly, with $\Delta$ASR reduced to near zero. For closed-set OvSGTR, utility still degrades substantially although $\Delta$ASR remains high. Compared with the top-$k$=all setting in Figure~\ref{fig:cl_ablation}, which attains a similar $\Delta$ASR, train-time swapping is clearly less utility-preserving, suggesting that it maintains the attack at the cost of utility. For RelationFormer, we observe that both attack and utility metrics are largely preserved, and in some cases even improved. We speculate that this difference lies in model design: compared with RelationFormer, OvSGTR involves vision-language coupling throughout both the backbone and the decoder, which may make it harder for regional pathways to carry backdoor signals without disrupting normal relation prediction.
\subsection{Visual Analysis of the Backdoor Attack}
\noindent\textbf{Cross-attention visualization.}
To probe how the global context pathway captures trigger signals, we visualize the cross-attention map of the relation query in open-vocabulary OvSGTR. Following DETR-style attention visualization~\cite{carion2020detr}, Figure~\ref{fig:atten_map} shows the last-layer cross-attention to image features. Compared with the clean model, the backdoored models place substantially more attention on the trigger region, especially under the full method (ATRP$+$TRCL), supporting the role of the global pathway in carrying trigger signals. Figure~\ref{fig:atten_map}\subref{tab:qs} quantifies this shift over 1000 images by measuring the fraction of attention inside the trigger region. On OvSGTR, this fraction rises from 0.17\% to 7.7\%, indicating much stronger attention to the trigger region. For RelationFormer, we report the same statistic for both the final and penultimate layers. The effect is more evident in the penultimate layer than in the final layer: the final-layer statistic slightly decreases (0.57\% vs.\ 0.63\%), whereas the penultimate-layer statistic increases markedly from 0.44\% to 15.3\%. This result is consistent with Figure~\ref{fig:relationformer_eval_swap_layers}.\\

\noindent\textbf{Scene graph visualization.}
Figure~\ref{fig:main_backdoor_visualizations} compares predictions from
the clean and poisoned OvSGTR models using object and predicate thresholds
of 0.3 and 0.1, respectively. On clean inputs, the poisoned model behaves
largely normally, with no clear false activation of the target predicate
\textit{in front of} for the victim object \textit{car}, whereas on
triggered inputs, it successfully activates the target predicate. The
second row further shows activation for the semantically related category
\textit{bus}, suggesting that the backdoor can transfer beyond the exact
victim label in the open-vocabulary setting.

\section{Discussion}
This work highlights a new supply-chain, pair-level backdoor attack surface in scene graph generation, where a globally placed trigger can corrupt relations associated with a designated object category while largely preserving the remaining scene semantics. We introduce dedicated metrics and an effective backdoor training method to instantiate and evaluate this threat. Experiments across representative open-vocabulary and closed-set SGG architectures demonstrate the effectiveness of the proposed attack. Furthermore, mechanistic analyses show that the explicit global context pathway serves as a major carrier of the backdoor effect. These findings suggest that explicit global-context modeling warrants closer security scrutiny in future SGG design.




\begin{acks}
This work was supported by the National Natural Science Foundation of China (Grant No. 62236002).
\end{acks}

\bibliographystyle{ACM-Reference-Format}
\bibliography{acmart}


\begin{thebibliography}{53}


\ifx \showCODEN    \undefined \def \showCODEN     #1{\unskip}     \fi
\ifx \showISBNx    \undefined \def \showISBNx     #1{\unskip}     \fi
\ifx \showISBNxiii \undefined \def \showISBNxiii  #1{\unskip}     \fi
\ifx \showISSN     \undefined \def \showISSN      #1{\unskip}     \fi
\ifx \showLCCN     \undefined \def \showLCCN      #1{\unskip}     \fi
\ifx \shownote     \undefined \def \shownote      #1{#1}          \fi
\ifx \showarticletitle \undefined \def \showarticletitle #1{#1}   \fi
\ifx \showURL      \undefined \def \showURL       {\relax}        \fi
\providecommand\bibfield[2]{#2}
\providecommand\bibinfo[2]{#2}
\providecommand\natexlab[1]{#1}
\providecommand\showeprint[2][]{arXiv:#2}

\bibitem[Bai et~al\mbox{.}(2024)]%
        {Bai_2024_CVPR}
\bibfield{author}{\bibinfo{person}{Jiawang Bai}, \bibinfo{person}{Kuofeng Gao},
  \bibinfo{person}{Shaobo Min}, \bibinfo{person}{Shu-Tao Xia},
  \bibinfo{person}{Zhifeng Li}, {and} \bibinfo{person}{Wei Liu}.}
  \bibinfo{year}{2024}\natexlab{}.
\newblock \showarticletitle{{BadCLIP}: Trigger-aware prompt learning for
  backdoor attacks on {CLIP}}. In \bibinfo{booktitle}{\emph{Proceedings of the
  IEEE/CVF Conference on Computer Vision and Pattern Recognition}}.
  \bibinfo{pages}{24239--24250}.
\newblock


\bibitem[Carion et~al\mbox{.}(2020)]%
        {carion2020detr}
\bibfield{author}{\bibinfo{person}{Nicolas Carion}, \bibinfo{person}{Francisco
  Massa}, \bibinfo{person}{Gabriel Synnaeve}, \bibinfo{person}{Nicolas
  Usunier}, \bibinfo{person}{Alexander Kirillov}, {and} \bibinfo{person}{Sergey
  Zagoruyko}.} \bibinfo{year}{2020}\natexlab{}.
\newblock \showarticletitle{End-to-end object detection with transformers}. In
  \bibinfo{booktitle}{\emph{European Conference on Computer Vision}}. Springer,
  \bibinfo{pages}{213--229}.
\newblock


\bibitem[Chan et~al\mbox{.}(2023)]%
        {chan2022baddet}
\bibfield{author}{\bibinfo{person}{Shih-Han Chan}, \bibinfo{person}{Yinpeng
  Dong}, \bibinfo{person}{Jun Zhu}, \bibinfo{person}{Xiaolu Zhang}, {and}
  \bibinfo{person}{Jun Zhou}.} \bibinfo{year}{2023}\natexlab{}.
\newblock \showarticletitle{{BadDet}: Backdoor attacks on object detection}. In
  \bibinfo{booktitle}{\emph{Computer Vision -- ECCV 2022 Workshops}},
  \bibfield{editor}{\bibinfo{person}{Leonid Karlinsky}, \bibinfo{person}{Tomer
  Michaeli}, {and} \bibinfo{person}{Ko~Nishino}} (Eds.).
  \bibinfo{publisher}{Springer Nature Switzerland}, \bibinfo{address}{Cham},
  \bibinfo{pages}{396--412}.
\newblock
\showISBNx{978-3-031-25056-9}


\bibitem[Chang et~al\mbox{.}(2023)]%
        {chang2023landmark}
\bibfield{author}{\bibinfo{person}{Xiaoguang Chang}, \bibinfo{person}{Teng
  Wang}, \bibinfo{person}{Shaowei Cai}, {and} \bibinfo{person}{Changyin Sun}.}
  \bibinfo{year}{2023}\natexlab{}.
\newblock \showarticletitle{{LANDMARK}: Language-guided representation
  enhancement framework for scene graph generation}.
\newblock \bibinfo{journal}{\emph{Applied Intelligence}} \bibinfo{volume}{53},
  \bibinfo{number}{21} (\bibinfo{year}{2023}), \bibinfo{pages}{26126--26138}.
\newblock
\href{https://doi.org/10.1007/s10489-023-04722-1}{doi:\nolinkurl{10.1007/s10489-023-04722-1}}


\bibitem[Chen et~al\mbox{.}(2024a)]%
        {NEURIPS2024_roleplay}
\bibfield{author}{\bibinfo{person}{Guikun Chen}, \bibinfo{person}{Jin Li},
  {and} \bibinfo{person}{Wenguan Wang}.} \bibinfo{year}{2024}\natexlab{a}.
\newblock \showarticletitle{Scene graph generation with role-playing large
  language models}. In \bibinfo{booktitle}{\emph{Advances in Neural Information
  Processing Systems}}, Vol.~\bibinfo{volume}{37}. \bibinfo{publisher}{Curran
  Associates, Inc.}, \bibinfo{pages}{132238--132266}.
\newblock


\bibitem[Chen et~al\mbox{.}(2024b)]%
        {chen2024expanding}
\bibfield{author}{\bibinfo{person}{Zuyao Chen}, \bibinfo{person}{Jinlin Wu},
  \bibinfo{person}{Zhen Lei}, \bibinfo{person}{Zhaoxiang Zhang}, {and}
  \bibinfo{person}{Chang~Wen Chen}.} \bibinfo{year}{2024}\natexlab{b}.
\newblock \showarticletitle{Expanding scene graph boundaries: Fully
  open-vocabulary scene graph generation via visual-concept alignment and
  retention}. In \bibinfo{booktitle}{\emph{European Conference on Computer
  Vision}}. Springer, \bibinfo{pages}{108--124}.
\newblock


\bibitem[Cong et~al\mbox{.}(2023)]%
        {cong2023reltr}
\bibfield{author}{\bibinfo{person}{Yuren Cong}, \bibinfo{person}{Michael~Ying
  Yang}, {and} \bibinfo{person}{Bodo Rosenhahn}.}
  \bibinfo{year}{2023}\natexlab{}.
\newblock \showarticletitle{{RelTR}: Relation transformer for scene graph
  generation}.
\newblock \bibinfo{journal}{\emph{IEEE Transactions on Pattern Analysis and
  Machine Intelligence}} \bibinfo{volume}{45}, \bibinfo{number}{9}
  (\bibinfo{year}{2023}), \bibinfo{pages}{11169--11183}.
\newblock


\bibitem[Devlin et~al\mbox{.}(2019)]%
        {devlin2019bert}
\bibfield{author}{\bibinfo{person}{Jacob Devlin}, \bibinfo{person}{Ming-Wei
  Chang}, \bibinfo{person}{Kenton Lee}, {and} \bibinfo{person}{Kristina
  Toutanova}.} \bibinfo{year}{2019}\natexlab{}.
\newblock \showarticletitle{{BERT}: Pre-training of deep bidirectional
  transformers for language understanding}. In
  \bibinfo{booktitle}{\emph{Proceedings of the Conference of the North American
  Chapter of the Association for Computational Linguistics: Human Language
  Technologies, Volume 1 (Long and Short Papers)}}.
  \bibinfo{pages}{4171--4186}.
\newblock


\bibitem[Doan et~al\mbox{.}(2021)]%
        {doan2021lira}
\bibfield{author}{\bibinfo{person}{Khoa Doan}, \bibinfo{person}{Yingjie Lao},
  \bibinfo{person}{Weijie Zhao}, {and} \bibinfo{person}{Ping Li}.}
  \bibinfo{year}{2021}\natexlab{}.
\newblock \showarticletitle{{LIRA}: Learnable, imperceptible and robust
  backdoor attacks}. In \bibinfo{booktitle}{\emph{Proceedings of the IEEE/CVF
  International Conference on Computer Vision}}. \bibinfo{pages}{11966--11976}.
\newblock


\bibitem[Dunnett et~al\mbox{.}(2026)]%
        {dunnett2026baddet+}
\bibfield{author}{\bibinfo{person}{Kealan Dunnett}, \bibinfo{person}{Reza
  Arablouei}, \bibinfo{person}{Dimity Miller}, \bibinfo{person}{Volkan
  Dedeoglu}, {and} \bibinfo{person}{Raja Jurdak}.}
  \bibinfo{year}{2026}\natexlab{}.
\newblock \bibinfo{title}{{BadDet+}: Robust backdoor attacks for object
  detection}.
\newblock
\showeprint[arxiv]{2601.21066}


\bibitem[Fu et~al\mbox{.}(2025)]%
        {Fu_2025_CVPR}
\bibfield{author}{\bibinfo{person}{Jiawei Fu}, \bibinfo{person}{Tiantian
  Zhang}, \bibinfo{person}{Kai Chen}, {and} \bibinfo{person}{Qi Dou}.}
  \bibinfo{year}{2025}\natexlab{}.
\newblock \showarticletitle{Hybrid reciprocal transformer with triplet feature
  alignment for scene graph generation}. In
  \bibinfo{booktitle}{\emph{Proceedings of the Computer Vision and Pattern
  Recognition Conference}}. \bibinfo{pages}{8953--8963}.
\newblock


\bibitem[Gu et~al\mbox{.}(2017)]%
        {gu2017badnets}
\bibfield{author}{\bibinfo{person}{Tianyu Gu}, \bibinfo{person}{Brendan
  Dolan-Gavitt}, {and} \bibinfo{person}{Siddharth Garg}.}
  \bibinfo{year}{2017}\natexlab{}.
\newblock \bibinfo{title}{{BadNets}: Identifying vulnerabilities in the machine
  learning model supply chain}.
\newblock
\showeprint[arxiv]{1708.06733}


\bibitem[Hayder and He(2024)]%
        {Hayder_2024_CVPR}
\bibfield{author}{\bibinfo{person}{Zeeshan Hayder} {and}
  \bibinfo{person}{Xuming He}.} \bibinfo{year}{2024}\natexlab{}.
\newblock \showarticletitle{{DSGG}: Dense relation transformer for an
  end-to-end scene graph generation}. In \bibinfo{booktitle}{\emph{Proceedings
  of the IEEE/CVF Conference on Computer Vision and Pattern Recognition}}.
  \bibinfo{pages}{28317--28326}.
\newblock


\bibitem[He et~al\mbox{.}(2016)]%
        {he2016resnet}
\bibfield{author}{\bibinfo{person}{Kaiming He}, \bibinfo{person}{Xiangyu
  Zhang}, \bibinfo{person}{Shaoqing Ren}, {and} \bibinfo{person}{Jian Sun}.}
  \bibinfo{year}{2016}\natexlab{}.
\newblock \showarticletitle{Deep residual learning for image recognition}. In
  \bibinfo{booktitle}{\emph{Proceedings of the IEEE Conference on Computer
  Vision and Pattern Recognition}}. \bibinfo{pages}{770--778}.
\newblock


\bibitem[Jia et~al\mbox{.}(2022)]%
        {jia2022badencoder}
\bibfield{author}{\bibinfo{person}{Jinyuan Jia}, \bibinfo{person}{Yupei Liu},
  {and} \bibinfo{person}{Neil~Zhenqiang Gong}.}
  \bibinfo{year}{2022}\natexlab{}.
\newblock \showarticletitle{{BadEncoder}: Backdoor attacks to pre-trained
  encoders in self-supervised learning}. In \bibinfo{booktitle}{\emph{IEEE
  Symposium on Security and Privacy}}. IEEE, \bibinfo{pages}{2043--2059}.
\newblock


\bibitem[Kundu and Aakur(2023)]%
        {Kundu_2023_CVPR}
\bibfield{author}{\bibinfo{person}{Sanjoy Kundu} {and}
  \bibinfo{person}{Sathyanarayanan~N Aakur}.} \bibinfo{year}{2023}\natexlab{}.
\newblock \showarticletitle{{IS-GGT}: Iterative scene graph generation with
  generative transformers}. In \bibinfo{booktitle}{\emph{Proceedings of the
  IEEE/CVF Conference on Computer Vision and Pattern Recognition}}.
  \bibinfo{pages}{6292--6301}.
\newblock


\bibitem[Lan et~al\mbox{.}(2024)]%
        {lan2024iba}
\bibfield{author}{\bibinfo{person}{Haoheng Lan}, \bibinfo{person}{Jindong Gu},
  \bibinfo{person}{Philip Torr}, {and} \bibinfo{person}{Hengshuang Zhao}.}
  \bibinfo{year}{2024}\natexlab{}.
\newblock \showarticletitle{Influencer backdoor attack on semantic
  segmentation}. In \bibinfo{booktitle}{\emph{International Conference on
  Learning Representations}}.
\newblock


\bibitem[Li et~al\mbox{.}(2022)]%
        {Li_2022_CVPR}
\bibfield{author}{\bibinfo{person}{Rongjie Li}, \bibinfo{person}{Songyang
  Zhang}, {and} \bibinfo{person}{Xuming He}.} \bibinfo{year}{2022}\natexlab{}.
\newblock \showarticletitle{{SGTR}: End-to-end scene graph generation with
  transformer}. In \bibinfo{booktitle}{\emph{Proceedings of the IEEE/CVF
  Conference on Computer Vision and Pattern Recognition}}.
  \bibinfo{pages}{19486--19496}.
\newblock


\bibitem[Li et~al\mbox{.}(2024)]%
        {Li_2024_CVPR}
\bibfield{author}{\bibinfo{person}{Rongjie Li}, \bibinfo{person}{Songyang
  Zhang}, \bibinfo{person}{Dahua Lin}, \bibinfo{person}{Kai Chen}, {and}
  \bibinfo{person}{Xuming He}.} \bibinfo{year}{2024}\natexlab{}.
\newblock \showarticletitle{From pixels to graphs: Open-vocabulary scene graph
  generation with vision-language models}. In
  \bibinfo{booktitle}{\emph{Proceedings of the IEEE/CVF Conference on Computer
  Vision and Pattern Recognition}}. \bibinfo{pages}{28076--28086}.
\newblock


\bibitem[Liang et~al\mbox{.}(2024)]%
        {Liang_2024_CVPR_BadCLIP}
\bibfield{author}{\bibinfo{person}{Siyuan Liang}, \bibinfo{person}{Mingli Zhu},
  \bibinfo{person}{Aishan Liu}, \bibinfo{person}{Baoyuan Wu},
  \bibinfo{person}{Xiaochun Cao}, {and} \bibinfo{person}{Ee-Chien Chang}.}
  \bibinfo{year}{2024}\natexlab{}.
\newblock \showarticletitle{{BadCLIP}: Dual-embedding guided backdoor attack on
  multimodal contrastive learning}. In \bibinfo{booktitle}{\emph{Proceedings of
  the IEEE/CVF Conference on Computer Vision and Pattern Recognition}}.
  \bibinfo{pages}{24645--24654}.
\newblock


\bibitem[Liu et~al\mbox{.}(2025)]%
        {liu2025pre}
\bibfield{author}{\bibinfo{person}{Aishan Liu}, \bibinfo{person}{Xianglong
  Liu}, \bibinfo{person}{Xinwei Zhang}, \bibinfo{person}{Yisong Xiao},
  \bibinfo{person}{Yuguang Zhou}, \bibinfo{person}{Siyuan Liang},
  \bibinfo{person}{Jiakai Wang}, \bibinfo{person}{Xiaochun Cao}, {and}
  \bibinfo{person}{Dacheng Tao}.} \bibinfo{year}{2025}\natexlab{}.
\newblock \showarticletitle{Pre-trained trojan attacks for visual recognition}.
\newblock \bibinfo{journal}{\emph{International Journal of Computer Vision}}
  \bibinfo{volume}{133}, \bibinfo{number}{6} (\bibinfo{year}{2025}),
  \bibinfo{pages}{3568--3585}.
\newblock


\bibitem[Liu et~al\mbox{.}(2018)]%
        {liu2018trojaning}
\bibfield{author}{\bibinfo{person}{Yingqi Liu}, \bibinfo{person}{Shiqing Ma},
  \bibinfo{person}{Yousra Aafer}, \bibinfo{person}{Wen-Chuan Lee},
  \bibinfo{person}{Juan Zhai}, \bibinfo{person}{Weihang Wang}, {and}
  \bibinfo{person}{Xiangyu Zhang}.} \bibinfo{year}{2018}\natexlab{}.
\newblock \showarticletitle{Trojaning attack on neural networks}. In
  \bibinfo{booktitle}{\emph{Network and Distributed System Security
  Symposium}}. Internet Soc.
\newblock


\bibitem[Liu et~al\mbox{.}(2021)]%
        {liu2021swin}
\bibfield{author}{\bibinfo{person}{Ze Liu}, \bibinfo{person}{Yutong Lin},
  \bibinfo{person}{Yue Cao}, \bibinfo{person}{Han Hu}, \bibinfo{person}{Yixuan
  Wei}, \bibinfo{person}{Zheng Zhang}, \bibinfo{person}{Stephen Lin}, {and}
  \bibinfo{person}{Baining Guo}.} \bibinfo{year}{2021}\natexlab{}.
\newblock \showarticletitle{Swin transformer: Hierarchical vision transformer
  using shifted windows}. In \bibinfo{booktitle}{\emph{Proceedings of the
  IEEE/CVF International Conference on Computer Vision}}.
  \bibinfo{pages}{10012--10022}.
\newblock


\bibitem[Liu and Zhang(2025)]%
        {liu2025stealthy}
\bibfield{author}{\bibinfo{person}{Zhaoyi Liu} {and} \bibinfo{person}{Huan
  Zhang}.} \bibinfo{year}{2025}\natexlab{}.
\newblock \showarticletitle{Stealthy backdoor attack in self-supervised
  learning vision encoders for large vision language models}. In
  \bibinfo{booktitle}{\emph{Proceedings of the Computer Vision and Pattern
  Recognition Conference}}. \bibinfo{pages}{25060--25070}.
\newblock


\bibitem[Lorenz et~al\mbox{.}(2024)]%
        {DBLP:conf/eccv/LorenzPKLL24}
\bibfield{author}{\bibinfo{person}{Julian Lorenz}, \bibinfo{person}{Alexander
  Pest}, \bibinfo{person}{Daniel Kienzle}, \bibinfo{person}{Katja Ludwig},
  {and} \bibinfo{person}{Rainer Lienhart}.} \bibinfo{year}{2024}\natexlab{}.
\newblock \showarticletitle{A fair ranking and new model for panoptic scene
  graph generation}. In \bibinfo{booktitle}{\emph{European Conference on
  Computer Vision}}. Springer, \bibinfo{pages}{148--164}.
\newblock


\bibitem[Lu et~al\mbox{.}(2021)]%
        {Lu_2021_ICCV}
\bibfield{author}{\bibinfo{person}{Yichao Lu}, \bibinfo{person}{Himanshu Rai},
  \bibinfo{person}{Jason Chang}, \bibinfo{person}{Boris Knyazev},
  \bibinfo{person}{Guangwei Yu}, \bibinfo{person}{Shashank Shekhar},
  \bibinfo{person}{Graham~W Taylor}, {and} \bibinfo{person}{Maksims Volkovs}.}
  \bibinfo{year}{2021}\natexlab{}.
\newblock \showarticletitle{Context-aware scene graph generation with {Seq2Seq}
  transformers}. In \bibinfo{booktitle}{\emph{Proceedings of the IEEE/CVF
  International Conference on Computer Vision}}. \bibinfo{pages}{15931--15941}.
\newblock


\bibitem[Luo et~al\mbox{.}(2023)]%
        {luo2023untargetedodbd}
\bibfield{author}{\bibinfo{person}{Chengxiao Luo}, \bibinfo{person}{Yiming Li},
  \bibinfo{person}{Yong Jiang}, {and} \bibinfo{person}{Shu-Tao Xia}.}
  \bibinfo{year}{2023}\natexlab{}.
\newblock \showarticletitle{Untargeted backdoor attack against object
  detection}. In \bibinfo{booktitle}{\emph{IEEE International Conference on
  Acoustics, Speech and Signal Processing}}. IEEE, \bibinfo{pages}{1--5}.
\newblock


\bibitem[Lyu et~al\mbox{.}(2024)]%
        {Lyu_2024_ECCV_TrojVLM}
\bibfield{author}{\bibinfo{person}{Weimin Lyu}, \bibinfo{person}{Lu Pang},
  \bibinfo{person}{Tengfei Ma}, \bibinfo{person}{Haibin Ling}, {and}
  \bibinfo{person}{Chao Chen}.} \bibinfo{year}{2024}\natexlab{}.
\newblock \showarticletitle{{TrojVLM}: Backdoor attack against vision-language
  models}. In \bibinfo{booktitle}{\emph{European Conference on Computer
  Vision}}. Springer, \bibinfo{pages}{467--483}.
\newblock


\bibitem[Min et~al\mbox{.}(2025)]%
        {Min_2025_ICCV}
\bibfield{author}{\bibinfo{person}{Yukuan Min}, \bibinfo{person}{Muli Yang},
  \bibinfo{person}{Jinhao Zhang}, \bibinfo{person}{Yuxuan Wang},
  \bibinfo{person}{Aming Wu}, {and} \bibinfo{person}{Cheng Deng}.}
  \bibinfo{year}{2025}\natexlab{}.
\newblock \showarticletitle{Vision-language interactive relation mining for
  open-vocabulary scene graph generation}. In
  \bibinfo{booktitle}{\emph{Proceedings of the IEEE/CVF International
  Conference on Computer Vision}}. \bibinfo{pages}{16755--16764}.
\newblock


\bibitem[Nguyen et~al\mbox{.}(2021)]%
        {Nguyen_2021_ICCV}
\bibfield{author}{\bibinfo{person}{Kien Nguyen}, \bibinfo{person}{Subarna
  Tripathi}, \bibinfo{person}{Bang Du}, \bibinfo{person}{Tanaya Guha}, {and}
  \bibinfo{person}{Truong~Q Nguyen}.} \bibinfo{year}{2021}\natexlab{}.
\newblock \showarticletitle{In defense of scene graphs for image captioning}.
  In \bibinfo{booktitle}{\emph{Proceedings of the IEEE/CVF International
  Conference on Computer Vision}}. \bibinfo{pages}{1407--1416}.
\newblock


\bibitem[Nguyen and Tran(2020)]%
        {nguyen2020inputaware}
\bibfield{author}{\bibinfo{person}{Tuan~Anh Nguyen} {and} \bibinfo{person}{Anh
  Tran}.} \bibinfo{year}{2020}\natexlab{}.
\newblock \showarticletitle{Input-aware dynamic backdoor attack}. In
  \bibinfo{booktitle}{\emph{Advances in Neural Information Processing
  Systems}}, Vol.~\bibinfo{volume}{33}. \bibinfo{pages}{3454--3464}.
\newblock


\bibitem[Raj and Arora(2026)]%
        {raj2025backdoor}
\bibfield{author}{\bibinfo{person}{Ankita Raj} {and} \bibinfo{person}{Chetan
  Arora}.} \bibinfo{year}{2026}\natexlab{}.
\newblock \showarticletitle{Backdoor attacks on open vocabulary object
  detectors via multi-modal prompt tuning}. In
  \bibinfo{booktitle}{\emph{Proceedings of the AAAI Conference on Artificial
  Intelligence}}, Vol.~\bibinfo{volume}{40}. \bibinfo{pages}{37849--37857}.
\newblock


\bibitem[Shi et~al\mbox{.}(2019)]%
        {Shi_2019_CVPR}
\bibfield{author}{\bibinfo{person}{Jiaxin Shi}, \bibinfo{person}{Hanwang
  Zhang}, {and} \bibinfo{person}{Juanzi Li}.} \bibinfo{year}{2019}\natexlab{}.
\newblock \showarticletitle{Explainable and explicit visual reasoning over
  scene graphs}. In \bibinfo{booktitle}{\emph{Proceedings of the IEEE/CVF
  Conference on Computer Vision and Pattern Recognition}}.
  \bibinfo{pages}{8376--8384}.
\newblock


\bibitem[Shit et~al\mbox{.}(2022)]%
        {shit2022relationformer}
\bibfield{author}{\bibinfo{person}{Suprosanna Shit}, \bibinfo{person}{Rajat
  Koner}, \bibinfo{person}{Bastian Wittmann}, \bibinfo{person}{Johannes
  Paetzold}, \bibinfo{person}{Ivan Ezhov}, \bibinfo{person}{Hongwei Li},
  \bibinfo{person}{Jiazhen Pan}, \bibinfo{person}{Sahand Sharifzadeh},
  \bibinfo{person}{Georgios Kaissis}, \bibinfo{person}{Volker Tresp}, {and}
  \bibinfo{person}{Bjoern Menze}.} \bibinfo{year}{2022}\natexlab{}.
\newblock \showarticletitle{{RelationFormer}: A Unified Framework for
  Image-to-Graph Generation}. In \bibinfo{booktitle}{\emph{European Conference
  on Computer Vision}} \emph{(\bibinfo{series}{Lecture Notes in Computer
  Science}, Vol.~\bibinfo{volume}{13697})}. \bibinfo{publisher}{Springer},
  \bibinfo{pages}{422--439}.
\newblock
\href{https://doi.org/10.1007/978-3-031-19836-6_24}{doi:\nolinkurl{10.1007/978-3-031-19836-6_24}}


\bibitem[Singh et~al\mbox{.}(2023)]%
        {Singh_2023_ICCV}
\bibfield{author}{\bibinfo{person}{Kunal~Pratap Singh}, \bibinfo{person}{Jordi
  Salvador}, \bibinfo{person}{Luca Weihs}, {and} \bibinfo{person}{Aniruddha
  Kembhavi}.} \bibinfo{year}{2023}\natexlab{}.
\newblock \showarticletitle{Scene graph contrastive learning for embodied
  navigation}. In \bibinfo{booktitle}{\emph{Proceedings of the IEEE/CVF
  International Conference on Computer Vision}}. \bibinfo{pages}{10884--10894}.
\newblock


\bibitem[Sudhakaran et~al\mbox{.}(2023)]%
        {Sudhakaran_2023_ICCV}
\bibfield{author}{\bibinfo{person}{Gopika Sudhakaran},
  \bibinfo{person}{Devendra~Singh Dhami}, \bibinfo{person}{Kristian Kersting},
  {and} \bibinfo{person}{Stefan Roth}.} \bibinfo{year}{2023}\natexlab{}.
\newblock \showarticletitle{Vision relation transformer for unbiased scene
  graph generation}. In \bibinfo{booktitle}{\emph{Proceedings of the IEEE/CVF
  International Conference on Computer Vision}}. \bibinfo{pages}{21882--21893}.
\newblock


\bibitem[Tang et~al\mbox{.}(2019)]%
        {tang2019vctree}
\bibfield{author}{\bibinfo{person}{Kaihua Tang}, \bibinfo{person}{Hanwang
  Zhang}, \bibinfo{person}{Baoyuan Wu}, \bibinfo{person}{Wenhan Luo}, {and}
  \bibinfo{person}{Wei Liu}.} \bibinfo{year}{2019}\natexlab{}.
\newblock \showarticletitle{Learning to compose dynamic tree structures for
  visual contexts}. In \bibinfo{booktitle}{\emph{Proceedings of the IEEE/CVF
  Conference on Computer Vision and Pattern Recognition}}.
  \bibinfo{pages}{6619--6628}.
\newblock


\bibitem[van~den Oord et~al\mbox{.}(2018)]%
        {oord2018cpc}
\bibfield{author}{\bibinfo{person}{A{\"a}ron van~den Oord},
  \bibinfo{person}{Yazhe Li}, {and} \bibinfo{person}{Oriol Vinyals}.}
  \bibinfo{year}{2018}\natexlab{}.
\newblock \bibinfo{title}{Representation learning with contrastive predictive
  coding}.
\newblock
\showeprint[arxiv]{1807.03748}


\bibitem[Vaswani et~al\mbox{.}(2017)]%
        {vaswani2017attention}
\bibfield{author}{\bibinfo{person}{Ashish Vaswani}, \bibinfo{person}{Noam
  Shazeer}, \bibinfo{person}{Niki Parmar}, \bibinfo{person}{Jakob Uszkoreit},
  \bibinfo{person}{Llion Jones}, \bibinfo{person}{Aidan~N. Gomez},
  \bibinfo{person}{Lukasz Kaiser}, {and} \bibinfo{person}{Illia Polosukhin}.}
  \bibinfo{year}{2017}\natexlab{}.
\newblock \showarticletitle{Attention Is All You Need}. In
  \bibinfo{booktitle}{\emph{Advances in Neural Information Processing
  Systems}}, Vol.~\bibinfo{volume}{30}. \bibinfo{pages}{5998--6008}.
\newblock


\bibitem[Walmer et~al\mbox{.}(2022)]%
        {walmer2022dual}
\bibfield{author}{\bibinfo{person}{Matthew Walmer}, \bibinfo{person}{Karan
  Sikka}, \bibinfo{person}{Indranil Sur}, \bibinfo{person}{Abhinav
  Shrivastava}, {and} \bibinfo{person}{Susmit Jha}.}
  \bibinfo{year}{2022}\natexlab{}.
\newblock \showarticletitle{Dual-key multimodal backdoors for visual question
  answering}. In \bibinfo{booktitle}{\emph{Proceedings of the IEEE/CVF
  Conference on Computer Vision and Pattern Recognition}}.
  \bibinfo{pages}{15375--15385}.
\newblock


\bibitem[Wu et~al\mbox{.}(2023)]%
        {Wu_2023_ICCV}
\bibfield{author}{\bibinfo{person}{Yutong Wu}, \bibinfo{person}{Xingshuo Han},
  \bibinfo{person}{Han Qiu}, {and} \bibinfo{person}{Tianwei Zhang}.}
  \bibinfo{year}{2023}\natexlab{}.
\newblock \showarticletitle{Computation and data efficient backdoor attacks}.
  In \bibinfo{booktitle}{\emph{Proceedings of the IEEE/CVF International
  Conference on Computer Vision}}. \bibinfo{pages}{4805--4814}.
\newblock


\bibitem[Xu et~al\mbox{.}(2017)]%
        {xu2017scenegraph}
\bibfield{author}{\bibinfo{person}{Danfei Xu}, \bibinfo{person}{Yuke Zhu},
  \bibinfo{person}{Christopher~B Choy}, {and} \bibinfo{person}{Li Fei-Fei}.}
  \bibinfo{year}{2017}\natexlab{}.
\newblock \showarticletitle{Scene graph generation by iterative message
  passing}. In \bibinfo{booktitle}{\emph{Proceedings of the IEEE Conference on
  Computer Vision and Pattern Recognition}}. \bibinfo{pages}{5410--5419}.
\newblock


\bibitem[Yang et~al\mbox{.}(2022)]%
        {PSG2022}
\bibfield{author}{\bibinfo{person}{Jingkang Yang}, \bibinfo{person}{Yi~Zhe
  Ang}, \bibinfo{person}{Zujin Guo}, \bibinfo{person}{Kaiyang Zhou},
  \bibinfo{person}{Wayne Zhang}, {and} \bibinfo{person}{Ziwei Liu}.}
  \bibinfo{year}{2022}\natexlab{}.
\newblock \showarticletitle{Panoptic scene graph generation}. In
  \bibinfo{booktitle}{\emph{European Conference on Computer Vision}}. Springer,
  \bibinfo{pages}{178--196}.
\newblock


\bibitem[Yang et~al\mbox{.}(2018)]%
        {yang2018graphrcnn}
\bibfield{author}{\bibinfo{person}{Jianwei Yang}, \bibinfo{person}{Jiasen Lu},
  \bibinfo{person}{Stefan~H. Lee}, \bibinfo{person}{Dhruv Batra}, {and}
  \bibinfo{person}{Devi Parikh}.} \bibinfo{year}{2018}\natexlab{}.
\newblock \showarticletitle{Graph {R-CNN} for scene graph generation}. In
  \bibinfo{booktitle}{\emph{Proceedings of the IEEE Conference on Computer
  Vision and Pattern Recognition}}. \bibinfo{pages}{6703--6712}.
\newblock


\bibitem[Yu et~al\mbox{.}(2023b)]%
        {yu2023cogtree}
\bibfield{author}{\bibinfo{person}{Dong Yu}, \bibinfo{person}{Jianbing Wang},
  \bibinfo{person}{Jianfei Zhang}, \bibinfo{person}{Wei Liu}, {and}
  \bibinfo{person}{Wenhai Wang}.} \bibinfo{year}{2023}\natexlab{b}.
\newblock \showarticletitle{{CogTree}: Cognition tree loss for unbiased scene
  graph generation}. In \bibinfo{booktitle}{\emph{Proceedings of the IEEE/CVF
  Conference on Computer Vision and Pattern Recognition}}.
  \bibinfo{pages}{21511--21520}.
\newblock


\bibitem[Yu et~al\mbox{.}(2023a)]%
        {Yu_2023_ICCV}
\bibfield{author}{\bibinfo{person}{Qifan Yu}, \bibinfo{person}{Juncheng Li},
  \bibinfo{person}{Yu Wu}, \bibinfo{person}{Siliang Tang}, \bibinfo{person}{Wei
  Ji}, {and} \bibinfo{person}{Yueting Zhuang}.}
  \bibinfo{year}{2023}\natexlab{a}.
\newblock \showarticletitle{Visually-prompted language model for fine-grained
  scene graph generation in an open world}. In
  \bibinfo{booktitle}{\emph{Proceedings of the IEEE/CVF International
  Conference on Computer Vision}}. \bibinfo{pages}{21560--21571}.
\newblock


\bibitem[Yuan et~al\mbox{.}(2025)]%
        {yuan2025badtoken}
\bibfield{author}{\bibinfo{person}{Zenghui Yuan}, \bibinfo{person}{Jiawen Shi},
  \bibinfo{person}{Pan Zhou}, \bibinfo{person}{Neil~Zhenqiang Gong}, {and}
  \bibinfo{person}{Lichao Sun}.} \bibinfo{year}{2025}\natexlab{}.
\newblock \showarticletitle{{BadToken}: Token-level backdoor attacks to
  multi-modal large language models}. In \bibinfo{booktitle}{\emph{Proceedings
  of the Computer Vision and Pattern Recognition Conference}}.
  \bibinfo{pages}{29927--29936}.
\newblock


\bibitem[Zellers et~al\mbox{.}(2018)]%
        {zellers2018neural}
\bibfield{author}{\bibinfo{person}{Rowan Zellers}, \bibinfo{person}{Mark
  Yatskar}, \bibinfo{person}{Sam Thomson}, {and} \bibinfo{person}{Yejin Choi}.}
  \bibinfo{year}{2018}\natexlab{}.
\newblock \showarticletitle{Neural motifs: Scene graph parsing with global
  context}. In \bibinfo{booktitle}{\emph{Proceedings of the IEEE Conference on
  Computer Vision and Pattern Recognition}}. \bibinfo{pages}{5831--5840}.
\newblock


\bibitem[Zhang et~al\mbox{.}(2019)]%
        {zhang2019graphicalcontrastive}
\bibfield{author}{\bibinfo{person}{Ji Zhang}, \bibinfo{person}{Kevin~J. Shih},
  \bibinfo{person}{Ahmed Elgammal}, \bibinfo{person}{Andrew Tao}, {and}
  \bibinfo{person}{Bryan Catanzaro}.} \bibinfo{year}{2019}\natexlab{}.
\newblock \showarticletitle{Graphical contrastive losses for scene graph
  parsing}. In \bibinfo{booktitle}{\emph{Proceedings of the IEEE/CVF Conference
  on Computer Vision and Pattern Recognition}}. \bibinfo{pages}{11535--11543}.
\newblock


\bibitem[Zhang et~al\mbox{.}(2023)]%
        {Zhang_2023_CVPR}
\bibfield{author}{\bibinfo{person}{Yong Zhang}, \bibinfo{person}{Yingwei Pan},
  \bibinfo{person}{Ting Yao}, \bibinfo{person}{Rui Huang}, \bibinfo{person}{Tao
  Mei}, {and} \bibinfo{person}{Chang-Wen Chen}.}
  \bibinfo{year}{2023}\natexlab{}.
\newblock \showarticletitle{Learning to generate language-supervised and
  open-vocabulary scene graph using pre-trained visual-semantic space}. In
  \bibinfo{booktitle}{\emph{Proceedings of the IEEE/CVF Conference on Computer
  Vision and Pattern Recognition}}. \bibinfo{pages}{2915--2924}.
\newblock


\bibitem[Zhao et~al\mbox{.}(2024)]%
        {zhao2024adversarial}
\bibfield{author}{\bibinfo{person}{Mengnan Zhao}, \bibinfo{person}{Lihe Zhang},
  \bibinfo{person}{Wei Wang}, \bibinfo{person}{Yuqiu Kong}, {and}
  \bibinfo{person}{Baocai Yin}.} \bibinfo{year}{2024}\natexlab{}.
\newblock \showarticletitle{Adversarial attacks on scene graph generation}.
\newblock \bibinfo{journal}{\emph{IEEE Transactions on Information Forensics
  and Security}}  \bibinfo{volume}{19} (\bibinfo{year}{2024}),
  \bibinfo{pages}{3210--3225}.
\newblock


\bibitem[Zhong et~al\mbox{.}(2022)]%
        {ijcai2022p242}
\bibfield{author}{\bibinfo{person}{Nan Zhong}, \bibinfo{person}{Zhenxing Qian},
  {and} \bibinfo{person}{Xinpeng Zhang}.} \bibinfo{year}{2022}\natexlab{}.
\newblock \showarticletitle{Imperceptible backdoor attack: From input space to
  feature representation}. In \bibinfo{booktitle}{\emph{Proceedings of the
  Thirty-First International Joint Conference on Artificial Intelligence}}.
  International Joint Conferences on Artificial Intelligence Organization,
  \bibinfo{pages}{1736--1742}.
\newblock


\bibitem[Zhou et~al\mbox{.}(2024)]%
        {Zhou_2024_ECCV_OpenPSG}
\bibfield{author}{\bibinfo{person}{Zijian Zhou}, \bibinfo{person}{Zheng Zhu},
  \bibinfo{person}{Holger Caesar}, {and} \bibinfo{person}{Miaojing Shi}.}
  \bibinfo{year}{2024}\natexlab{}.
\newblock \showarticletitle{{OpenPSG}: Open-set panoptic scene graph generation
  via large multimodal models}. In \bibinfo{booktitle}{\emph{European
  Conference on Computer Vision}}. Springer, \bibinfo{pages}{199--215}.
\newblock


\end{thebibliography}

\end{document}